\documentclass[12pt,nonatbib]{elsarticle}

\usepackage{tabularx}
\usepackage{booktabs}
\usepackage{float}
\usepackage{xcolor}
\usepackage{soul}

\usepackage{rotating}
\usepackage{multirow}

\usepackage{xcolor}

\usepackage{pdflscape}

\usepackage{pgfplots}
\pgfplotsset{compat=1.18}

\usepackage{fontawesome5} %
\usepackage[table]{xcolor}
\usepackage{tikz}
\usetikzlibrary{arrows.meta,calc}   %
\usepackage{booktabs}

\definecolor{okabeBlack}{HTML}{000000}
\definecolor{okabeOrange}{HTML}{E69F00}
\definecolor{okabeSkyBlue}{HTML}{56B4E9}
\definecolor{okabeBluishGreen}{HTML}{009E73}
\definecolor{okabeYellow}{HTML}{F0E442}
\definecolor{okabeBlue}{HTML}{0072B2}
\definecolor{okabeVermillion}{HTML}{D55E00}
\definecolor{okabeReddishPurple}{HTML}{CC79A7}
\definecolor{neutralGrey}{HTML}{999999}

\usepackage{amssymb}
\usepackage{amsmath}

\usepackage[colorlinks,
pdfpagelabels,
pdfstartview = FitH,
bookmarksopen = true,
bookmarksnumbered = true,
linkcolor = uhhstone,
plainpages = false,
hypertexnames = false,
citecolor = uhhstone]{hyperref} %
\usepackage[printonlyused,withpage]{acronym}

\usepackage{color,soul}
\usepackage[british]{babel}
\usepackage{csquotes}
\makeatletter\let\c@author\relax\makeatother
\usepackage[style=numeric-comp,sorting=none,sortcites=true,backend=biber]{biblatex}
\providecommand{\nptextcite}{\cite}
\DefineBibliographyExtras{british}{}
\AtBeginBibliography{\raggedright}
\usepackage{lineno}
\usepackage{orcidlink}
\usepackage{pgfplots}
\pgfplotsset{compat=1.18}
\usepackage{geometry}
\journal{Transportation Research Interdisciplinary Perspectives}

\makeatletter
\long\def\@makecaption#1#2{%
  \vskip\abovecaptionskip
  \sbox\@tempboxa{\textbf{#1.} #2}%
  \ifdim \wd\@tempboxa >\hsize
    \textbf{#1.} #2\par
  \else
    \global \@minipagefalse
    \hb@xt@\hsize{\hfil\box\@tempboxa\hfil}%
  \fi
  \vskip\belowcaptionskip}
\makeatother

\begin{document}

    \acrodef{ilu}[ILU]{Intermodal Loading Unit}
    \acrodef{tu}[TU]{transportation unit}
    \acrodef{uic}[UIC]{International Union of Railways}

    \acrodef{uav}[UAV]{unmanned aerial vehicle}

    \acrodef{cer}[CER]{character error rate}
    \acrodef{iou}[IoU]{intersection over union}
    \acrodef{map}[mAP]{mean average precision}
    \acrodef{ap}[AP]{average precision}
    \acrodef{fps}[FPS]{frames per second}
    \acrodef{gflop}[GFLOP]{giga floating-point operation}

    \acrodef{trudi}[TRUDI]{TRansportation Unit Detection and Identification}
    \acrodef{p-scr}[PRISMA-ScR]{Preferred Reporting Items for Systematic Reviews and Meta-Analyses extension for Scoping Reviews}

    \acrodef{pcc}[PCC]{Population–Concept–Context}

    \acrodef{dip}[DIP]{Digital Image Processing}
    \acrodef{ml}[ML]{Machine Learning}
    \acrodef{dl}[DL]{Deep Learning}

    \acrodef{cv}[CV]{Computer Vision}

    \acrodef{ocr}[OCR]{Optical Character Recognition}
    \acrodef{accr}[ACCR]{Automatic Container Code Recognition}

    \acrodef{svm}[SVM]{Support Vector Machine}
    \acrodef{mlp}[MLP]{Multi-Layer Perceptron}
    \acrodef{cnn}[CNN]{Convolutional Neural Network}
    \acrodef{gan}[GAN]{Generative Adversarial Network}
    \acrodef{sift}[SIFT]{Scale-Invariant Feature Transform}
    \acrodef{hog}[HOG]{Histogram of Oriented Gradients}

\newif\ifshowchanges

\ifshowchanges
  \newcommand{\hlone}[1]{{\sethlcolor{yellow}\hl{#1}}}
  \newcommand{\hltwo}[1]{{\sethlcolor{cyan!25}\hl{#1}}}
\else
  \newcommand{\hlone}[1]{#1}
  \newcommand{\hltwo}[1]{#1}
\fi

\soulregister{\cite}{7}
\soulregister{\ref}{7}
\soulregister{\parencite}{7}
\soulregister{\textcite}{7}
\soulregister{\nptextcite}{7}

\makeatletter
\@for\my@cmd:=ac,acs,acl,acf,acp,acsp,aclp,acfp,acfi,acsu,aclu,acused,%
              Ac,Acs,Acl,Acf,Acp,Acsp,Aclp,Acfp,Acfi,Acsu,Aclu,iac,Iac\do{%
  \expandafter\soulregister\csname\my@cmd\endcsname{7}%
}
\makeatother

\begin{frontmatter}

\title{Automatic Intermodal Loading Unit Identification \\ using Computer Vision: A Scoping Review}

\affiliation[cv]{organization={University of Hamburg},%
            addressline={Department of Informatics, Computer Vision Group},
            city={Hamburg},
            country={Germany}}

\affiliation[dos]{organization={University of Hamburg},%
            addressline={Department of Informatics, Distributed Operating Systems Group},
            city={Hamburg},
            country={Germany}}

\affiliation[tuhh]{organization={Hamburg University of Technology},%
            addressline={Institute of Maritime Logistics},
            city={Hamburg},
            country={Germany}}

\cortext[corresponding]{indicates corresponding author (emre.guelsoylu@uni-hamburg.de).}

\author[cv]{Emre Gülsoylu\orcidlink{0000-0002-3834-3645}\corref{corresponding}}
\author[dos]{Alhassan Abdelhalim\orcidlink{0009-0009-4523-447X}}
\author[tuhh]{Derya Kara Boztas\orcidlink{0009-0002-1793-4830}}
\author[tuhh]{Ole Grasse\orcidlink{0000-0003-1982-9436}}
\author[tuhh]{Carlos Jahn\orcidlink{0000-0002-5409-0748}}
\author[cv]{Simone Frintrop\orcidlink{0000-0002-9475-3593}}
\author[dos]{Janick Edinger\orcidlink{0000-0002-9392-2922}}

\begin{abstract}
    \noindent \textbf{Background:} The standardisation of Intermodal Loading Units (ILUs), including containers, semi-trailers, and swap bodies, has transformed global trade, yet efficient and robust identification remains an operational bottleneck in ports and terminals. \\
    \noindent \textbf{Objective:} To map Computer Vision (CV) methods for ILU identification, clarify terminology, summarise \hlone{the evolution of proposed approaches}, and highlight research gaps, future directions \hlone{and their potential effects on terminal operations}.\\
    \noindent \textbf{Methods:} Following PRISMA-ScR, we searched Google Scholar\hltwo{, dblp and OpenAlex} for English-language studies with quantitative results. After \hltwo{screening}, the studies were charted across methods, datasets, failure modes, and evaluation metrics.\\
    \noindent \textbf{Results:} \hltwo{We reviewed 95 empirical studies on CV-based solutions for the ILU identification task, published between 1990 and 2026.}
    Methodological evolution \hlone{of ILU identification solutions, datasets, evaluation of the proposed methods} and future research directions are summarised. A shift from static (\hltwo{e.g.,} OCR-gates) to \hltwo{vehicle-mounted} camera setups, which \hltwo{can enable} precise port monitoring\hltwo{,} is observed. \hltwo{Only 26 of the 95 studies report an end-to-end accuracy (median 94.5~\%, range 19--100~\%), which is relevant for the operations. Eight datasets are publicly available, yet none of them is commonly used as a benchmark dataset, and six studies describe a deployed system and seven a field test.} \\
    \noindent \textbf{Conclusions:} \hltwo{We propose the term ILU identification for this broader task, which includes finding the unit and its code and covers semi-trailers and swap bodies. Yet the literature still reflects its origin at the gate. The challenges that are typical of moving cameras receive the least attention, and no study reports separate results for semi-trailers or swap bodies. Evaluation requires public benchmarks that cover these new conditions and all ILU types, with end-to-end accuracy results. Promising methodological directions are synthetic data, unified end-to-end models and recognisers built for codes without linguistic context. Vehicle-mounted cameras also require mission planning with automatic coverage path planning as future work. Addressing these challenges has the potential to reduce human errors, turnaround times and operational costs and to increase resilience against disruptions. }
\end{abstract}

    \begin{keyword}
    \def\sep{\unskip; }
    \interlinepenalty=10000 %
    Intermodal loading unit identification \sep
    Container code recognition \sep
    Scene-text spotting \sep
    Computer vision \sep
    Scoping review
    \end{keyword}

\emergencystretch=3em %

    \begin{keyword}
    \def\sep{\unskip; }
    Intermodal loading unit identification \sep
    Container code recognition \sep
    Scene-text spotting \sep
    Computer vision \sep
    Scoping review

    \end{keyword}
\end{frontmatter}

\section{Introduction}
\label{sec:introduction}

The term \ac{ilu} has been introduced to describe units designed to be easily transferred between different modes of transport, such as vessels, trains, and trucks, without the need to unload and reload the contents. The loading units, such as containers, interchangeable bodies (swap bodies) and cranable semi-trailers, fall under this term \parencite{FilinaDawidowicz2022complexity}.
A unique identification code, following the \hltwo{ISO~6346} standard \parencite{ISO6346}, is assigned to each \ac{ilu} for tracking purposes. This code consists of \hltwo{a three-letter owner code, a one-letter equipment category identifier, a six-digit serial number and a check digit}. \hltwo{Identical-looking} \acp{ilu} can \hltwo{only} be distinguished by their ID codes.

While the \hlone{standardised ID codes provide a framework for tracking of} \acp{ilu}, \hlone{the scale of modern port operations introduces challenges}. As global trade expands, many leading ports handle tens of millions of containers annually, with some operating at or above their nominal design capacity \parencite{lange2017reducing}. In this high throughput, keeping track of \acp{ilu} is important to ensure efficient management \parencite{parola2017drivers}.
\hlone{For example, at transfer points of} the terminal to other modes of transport, each \ac{ilu} must be identified and cross-checked with its associated documentation before entry or exit. When automated systems fail to achieve reliable \ac{ilu} identification, manual verification becomes necessary, further slowing down the process. Even a brief slowdown at a busy gate can cause truck queues stretching hundreds of metres, disrupting yard operations and increasing turnaround times for transport vehicles \parencite{spasovic2015quantifying, bett2024simulation}. Extended waiting times contribute to congestion which increases fuel consumption and greenhouse gas emissions from idling trucks, amplifying the environmental footprint of port activities \parencite{giuliano2007reducing}. Therefore, \hlone{efficient and automatic \ac{ilu} identification }\hltwo{can help to overcome} these bottlenecks and \hltwo{to sustain} the growth of containerised transport.

Standardisation and digitalisation enabled the development of various techniques for \ac{ilu} identification. Among these, \ac{cv}-based \hltwo{\ac{ilu} identification is an} alternative to RFID, which \hltwo{at the checkpoints of one port still left 20--30~\% of the drivers to stop and show their documents} \parencite{tai2011method} and entails high installation and maintenance costs \parencite{shithil2022container, yu2024two, ngo2023vision, tran2026automatic, khuboni2025confidence}. In contrast, \ac{cv}-based methods can be implemented using a range of camera setups, from static \acused{ocr}\hltwo{Optical Character Recognition (\acs{ocr})-gates} \parencite{garmouch2025enhancing, filom2022applications}, \hltwo{which} can identify entering and exiting \acp{ilu} \parencite{li2022towards, zhao2024practical}, to more flexible options such as \hltwo{camera-equipped} aerial or ground vehicles, or even mobile devices, which \hltwo{can} reduce the need for specialised infrastructure \parencite{teegen2024drone, liu2025lightweight, gulsoylu2025trudi, gulsoylu2025inland, chen2026end, yu2024two}. Furthermore, \ac{cv}-based \ac{ilu} identification \hltwo{can contribute to} security by providing accurate verification for customs clearance, which can \hltwo{support the} correct declaration of goods \parencite{hlabisa2024automated, ngo2023vision, huang2013study, roeksukrungrueang2018implementation}.

Despite its \hltwo{potential to improve} \hlone{port operations by reducing human-related errors} and \hlone{increasing the processing speed} \parencite{roeksukrungrueang2018implementation, zhang2019deep, hsu2023automatic, khuboni2025confidence}, to the best of our knowledge, \hltwo{the only literature review on the topic is a narrative overview of container number recognition by} \textcite{wang2026comprehensive}\hltwo{. It treats RFID, video recognition and \ac{ocr} side by side and summarises preprocessing, localisation, segmentation and character-recognition methods from about thirty references, without stating how these were searched for or selected. The present review differs from it in four respects: (1) it follows a documented search and selection procedure with stated eligibility criteria, (2) it is restricted to \ac{cv}-based identification, and its eligibility criteria cover all \ac{ilu} types (containers, swap bodies and semi-trailers) instead of only containers, (3) it charts every included study on the same data items (datasets, camera positions, evaluation metrics, end-to-end results, computational efficiency and failure modes) and publishes the resulting chart, and (4) it addresses the inconsistent terminology of the field.

Broader reviews of port technologies and of machine learning in port and hinterland operations treat container code recognition as one application among many, without a focus on \ac{cv} solutions on the topic} \parencite{kosiek2021analysis, lukashevich2025machine, rodrigues2026advancing, tran2026smart}\hltwo{.} \hlone{While prior surveys have addressed similar industrial problems, such as automatic licence plate recognition} \parencite{shashirangana2020automated, lubna2021automatic, khan2023license, ismail2025automatic}\hlone{ or product label recognition} \parencite{santra2019comprehensive, guimaraes2023review}\hlone{, none have systematically reviewed intermodal loading unit identification.} \hltwo{\ac{ilu} identification differs from these tasks. The codes are read within terminal operations, at gates, under cranes and in the yard, where the handling equipment, the position of the code on the unit and the time available for a read are dictated by the operation. Moreover, the images are acquired from multiple perspectives (fixed gate and crane cameras, vehicle-mounted and aerial cameras).}

To fill this gap, we prepared this literature review by following \hltwo{the} \ac{p-scr} guidelines \parencite{tricco2018prisma} with the intention of identifying the available publications and proposed concepts. \hlone{This review is timely as the research is actively shifting away from static, infrastructure-heavy }\hltwo{``OCR-gates''}\hlone{ at terminal entrances towards dynamic, vehicle-mounted or aerial camera setups (e.g., on drones, reach stackers }\hltwo{and terminal trucks).}

\hlone{The motivation for this shift is operational rather than computational. Vehicle-mounted cameras traverse the yard during normal handling operations and therefore cover substantially more area and viewpoints than fixed gates or surveillance cameras. This enables \ac{cv} solutions to be applied }\hltwo{to}\hlone{ broader logistics operations, such as yard inventory updates}\hltwo{ and}\hlone{ in-transit identification, rather than only }\hltwo{at the gate. Such applications could improve yard inventory accuracy by enabling more frequent verification of \ac{ilu} locations, support tracking across terminal areas, and facilitate the verification of \acp{ilu} during transfers between vessels, trucks, and trains.}

\hlone{With this review, we aim to (1) clear up ambiguous terminology, (2) track the historical evolution of these technologies, (3) emphasise the importance of publicly available benchmark datasets, (4) point to inconsistent metrics, and (5) outline current research gaps.}

In summary, there are \hltwo{four} contributions of this review:
\begin{itemize}
  \item Catalogue \hltwo{the eligible} empirical studies on \ac{ilu} identification using \ac{cv} methods between 1990 and \hltwo{2026},
  \item Re-evaluate the terminology used in \hltwo{the} \ac{ilu} identification field and provide clear definitions,
  \item Synthesise an overview of \ac{ilu} identification methods, image acquisition settings, datasets, \hltwo{failure modes} and evaluation metrics \hltwo{reported} in these studies,
  \item Highlight research gaps and suggest future priorities.
\end{itemize}

\section{Methodology}\label{sec:methodology}
\begin{sloppypar}\hlone{This literature review was conducted in accordance with the \ac{p-scr} guidelines} \parencite{tricco2018prisma} \hlone{with four research questions:}\end{sloppypar}

\begin{enumerate}
    \item \hlone{What terminology has been used in the literature to describe automatic ILU identification, and how does this terminology constrain searchability, comparability, and scope?}
    \item \hlone{How have the dominant CV approaches for ILU identification evolved between 1990 and} \hltwo{2026}\hlone{, and what factors triggered each methodological shift?}
    \item \hlone{What datasets and evaluation metrics are reported in the literature, and how do these choices affect the comparability of published results?}
    \item \hlone{Which challenging conditions and operational settings remain under-addressed, and what research directions are most relevant for moving the field from} \hltwo{static OCR-gates} \hlone{towards dynamic, vehicle-mounted deployments?}
\end{enumerate}

Eligibility criteria were specified \textit{a priori} and piloted before the screening began. These criteria follow the \ac{pcc} logic and Table~\ref{tab:eligibility} details the inclusion and exclusion rules \hltwo{for each domain}.

\begin{table}
  \caption{Inclusion and exclusion criteria of this scoping review of \ac{ilu} identification.}
  \label{tab:eligibility}
  \renewcommand{\arraystretch}{1.15}
  \setlength{\tabcolsep}{4pt}
  \centering
  \small
  \newcolumntype{L}[1]{>{\raggedright\arraybackslash}m{#1}}
  \renewcommand{\tabularxcolumn}[1]{m{#1}}
  \begin{tabularx}{\textwidth}{@{}L{3cm} >{\raggedright\arraybackslash}X >{\raggedright\arraybackslash}X@{}}
    \toprule
    \textbf{Domain} & \textbf{Inclusion rule} & \textbf{Exclusion rule} \\
    \midrule
    Date range
      & \hltwo{1 January 1990}–\hltwo{12 October 2026}
      & $<$ 1990 or after cut-off \\[2pt]

    Language
      & English full text
      & Non-English; abstract-only \\[2pt]

    Publication type
      & Peer-reviewed journal / conference paper; pre-print (quantitative)
      & Patents, \hltwo{standards}, theses, \hltwo{project reports, datasets, software repositories,} news or editorials \\[2pt]

    Population (P)
      & ILUs (\hltwo{containers, swap bodies, semi-trailers})
      & Other objects unless ILUs also analysed \\[2pt]

    Concept (C)
      & CV localisation \emph{\hltwo{and/or}} recognition of alphanumeric ILU code
      & RFID/barcode solutions \\[2pt]

    Context (C)
      & Fixed or mobile cameras in ports, terminals, rail yards
      & \hltwo{Warehouses and indoor factory settings} \\[2pt]

    Reporting completeness
      & Method detail \emph{plus} numeric outcomes (precision, recall, F1, mAP, read-rate)
      & No quantitative results \\[2pt]

    Duplicates / accessibility
      & \hltwo{Peer-reviewed version if there is one, otherwise the first} complete, accessible full text
      & Duplicates; inaccessible full text \\
    \bottomrule
  \end{tabularx}
  \renewcommand{\arraystretch}{1}%
\end{table}

As the first \hltwo{scoping} review on \ac{ilu} identification, we applied inclusive eligibility criteria to capture the full scope of available evidence. Our review includes articles written in English, published from 1990 to \hltwo{12 October 2026}, that follow a scientific structure. Both peer-reviewed articles and preprints are included as long as they report quantitative results, allowing us to catalogue the latest research findings.

We \hltwo{searched} Google Scholar\hltwo{,} dblp-Computer Science Bibliography \hltwo{and OpenAlex up to 12 October 2026} to find scientific articles on \ac{ilu} identification using \ac{cv} or \ac{dip} techniques. \hlone{Our search combined a high-recall academic search engine (Google Scholar) to cover even small venues, and a high-precision computer science-focused index (dblp). The choice of databases was deliberate and follows from three considerations: (1) the nature of a scoping review, (2) the terminological instability of this field, and (3) the distribution of venues that publish ILU-identification work.} \hltwo{OpenAlex, an open index whose results can be exported reproducibly, complements the two.}

Our search terms included \textit{container code recognition}, \hlone{\textit{container number recognition}, \textit{container identification}, \textit{BIC code recognition}, \textit{intermodal loading unit code recognition}}\hltwo{,}\hlone{ \textit{trailer code recognition}, and \textit{transportation unit identification}}. As discussed in Section~\ref{sec:terminology_beyon_ocr_gates}, the terminology used in this field is not standardised and it directly affects our search methodology. We checked the cited papers and \hltwo{`related articles'} in Google Scholar and Research Rabbit\footnote{\href{https://www.researchrabbit.ai/}{https://www.researchrabbit.ai/}, accessed 29 September 2026.} to expand our pool of eligible articles.

\hltwo{We included further phrases found in eligible studies, such as \textit{container ISO code}, \textit{container identity code}, \textit{container marking recognition} and \textit{ISO6346}. On Google Scholar, up to the first 200 results of each string were screened. On dblp, 16 terms were searched. As Google Scholar does not provide reproducible export, we also searched OpenAlex, an open bibliographic database that indexes titles, abstracts and full texts, with the same strings.

Supplementary Table~S4 lists every string with its source, date and number of results. Duplicates across the sources were removed, and records outside the date range, not in English, of an excluded publication type (theses, project reports, patents, standards, datasets, software repositories) or off-topic were excluded by keyword rules: a record counted as off-topic if none of its title, snippet and abstract mentions an \ac{ilu} term such as \textit{container} or \textit{trailer} combined with a code term such as \textit{number}, \textit{code} or \textit{marking}, or a term such as \textit{ISO~6346} or \textit{BIC code}, or if its title places it in another field, such as software containers, IT security or medicine. The remaining records were screened by title, snippet and abstract. The full texts of those that passed were assessed against the eligibility criteria (Table~\ref{tab:eligibility}). Two authors shared this screening. Disagreements on the final selection and on the data charting were resolved by consensus among the authors}.

From the \hltwo{95} eligible articles\hltwo{,} the \hltwo{data items listed in Supplementary Section~S1} were extracted into a spreadsheet\footnote{Available in supplementary materials.}\hltwo{.} \hltwo{Every extracted item is backed in the chart by a quote of the paper. Two authors checked every row of the chart.} The resulting table was then analysed and discussed by the authors.

\section{Results and Discussion}
\hltwo{After the study selection and a descriptive bibliometric overview, each research question is answered: RQ1 in Section~\ref{sec:terminology_beyon_ocr_gates}, RQ2 in Section~\ref{sec:methods}, RQ3 in Sections~\ref{sec:datasets} and~\ref{sec:evaluation} and RQ4 in Sections~\ref{sec:failure_modes} and~\ref{sec:future_directions}.} Section~\ref{sec:limitations} \hltwo{discusses the limitations of the review.}

\hltwo{The search identified 2,925 records. After 91 duplicates and 2,130 records excluded by rule (1,695 off-topic by keyword rules and 435 outside the date range, not in English or of an excluded publication type) had been removed, 704 records were screened. Of these, 543 were excluded on title and abstract, 161 full texts were sought, 30 of them could not be retrieved, and 131 were assessed for eligibility. At full text, articles were excluded for focusing on a task other than \ac{ilu} identification via ISO~6346-compliant codes (e.g., container detection, damage assessment or other industrial text recognition like wagon code recognition) ($n=16$), for not reporting quantitative results ($n=9$), for being written in a non-English language or being patents, project reports, dissertations ($n=7$), or as duplicates or concept papers ($n=4$).} \hltwo{The first of these criteria also excludes studies on neighbouring \ac{cv} tasks that are relevant to \ac{ilu} identification but neither localise nor read the code, for example the classification of transport-vehicle images into articulated trucks, trailers, containers and tank containers} \parencite{duranobregon2026comparison}\hltwo{. In total, 95 studies were included and charted.} Figure~\ref{fig:fig_prisma_flow} \hlone{illustrates the flow for obtaining eligible articles.}

\begin{figure}
    \centering
\begin{tikzpicture}[
    font=\scriptsize, >={Stealth[length=4pt]}, line width=0.4pt,
    box/.style={draw=okabeOrange, line width=0.6pt, rounded corners=1.5pt, align=center, inner sep=2.5pt, fill=white, anchor=north west, text width=3.6cm},
    side/.style={draw=okabeOrange, line width=0.6pt, rounded corners=1.5pt, align=left, inner sep=2.5pt, fill=white, anchor=north west, text width=7.4cm},
    head/.style={draw=okabeOrange, fill=okabeOrange!35, line width=0.6pt, rounded corners=1.5pt, align=center, inner sep=2.5pt, anchor=north west, text width=3.6cm, font=\scriptsize\bfseries},
    phase/.style={rotate=90, font=\scriptsize\bfseries, anchor=center},
    arr/.style={->, draw=black},
    every node/.append style={execute at begin node={\hyphenpenalty=10000\exhyphenpenalty=10000\relpenalty=10000\binoppenalty=10000\thickmuskip=5mu\medmuskip=4mu}}]
  \def\xM{0.45}\def\xS{4.85}   %
  \node[box]  (M1) at (\xM,0)     {Records identified in Google~Scholar, dblp and OpenAlex and by citation searching ($n = 2{,}925$)};
  \node[side] (S1) at (\xS,0)     {Records removed before screening ($n = 2{,}221$):\\
      off-topic by keyword rules 1,695\\ language, date range or publication type 435\\ duplicates 91};
  \node[box]  (M2) at (\xM,-1.85) {Records screened ($n = 704$)};
  \node[side] (S2) at (\xS,-1.85) {Records excluded on title and abstract ($n = 543$)};
  \node[box]  (M3) at (\xM,-2.8) {Records sought for retrieval ($n = 161$)};
  \node[side] (S3) at (\xS,-2.8) {Records not retrieved ($n = 30$)};
  \node[box]  (M4) at (\xM,-4.05) {Records assessed for eligibility ($n = 131$)};
  \node[side] (S4) at (\xS,-4.05) {Records excluded ($n = 36$):\\
      other task, e.g.\ container detection or wagon codes 16\\ no quantitative results for ILU codes 9\\ non-English or patent 7\\ duplicates and concept papers 4};
  \node[head] (M5) at (\xM,-6.05)  {Studies included in the review ($n = 95$)};
  \foreach \i/\j in {1/2,2/3,3/4,4/5} {\draw[arr] (M\i) -- (M\j);}
  \foreach \i in {1,...,4} {\draw[arr] (M\i.east) -- (M\i.east -| S\i.west);}
  \node[phase] at (0.1,-0.65) {Identification};
  \node[phase] at (0.1,-3.65) {Screening};
  \node[phase] at (0.1,-6.35) {Included};
\end{tikzpicture}
    \caption{\hlone{PRISMA-ScR flow diagram of the study selection process.} \hltwo{Sources, search strings and exclusion rules are described in Section~\ref{sec:methodology}.}}
    \label{fig:fig_prisma_flow}
\end{figure}

\begin{figure}
    \centering
    \includegraphics[width=\linewidth]{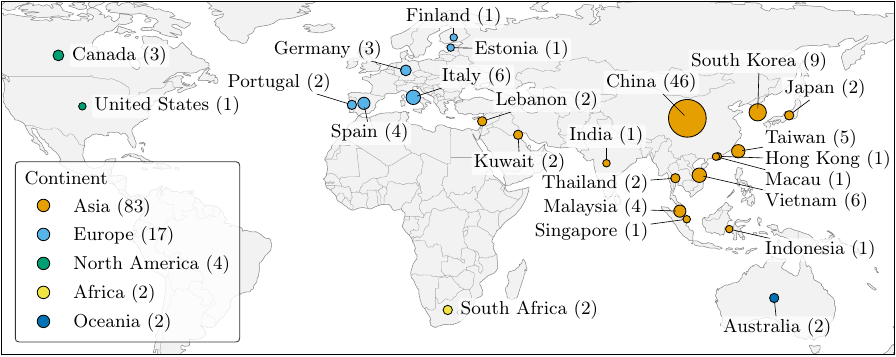}
    \caption{Geographic distribution of articles by country (n=\hltwo{95}). \hltwo{Ten articles have authors from more than one country or region, which adds 13 mentions (108 in total for 95 articles).} \hltwo{The area of a marker is proportional to the number of articles, its colour shows the continent.}}
    \label{fig:publications_world_map}
\end{figure}

\hltwo{The following bibliometric overview is descriptive: it shows where the studies come from, how they were funded and in which outlets they appeared.} Most of the published articles were written by authors with Asian affiliations (\hltwo{76.85~\%} \hltwo{of the country mentions}) as shown in Figure~\ref{fig:publications_world_map}. \hltwo{This concentration may reflect} the high volumes of \ac{ilu} handling in Asian ports and terminals, which creates a need for efficient \ac{ilu} identification systems. Countries such as China, Taiwan, South Korea, Vietnam, and Japan play significant roles in global trade. \hltwo{This may encourage} innovation\hltwo{,} as reflected by the contribution of Asian countries to the scientific publications. Substantial investment in research and development by Asian governments and companies further supports this innovation \parencite{unescap2024, liu2025green}.

Public funding is \hltwo{the main source among the 54 studies that report their funding (public funding alone: 47 studies, 49~\% of all 95 publications; industry funding or public-industry collaborations: six; no external funding: one; 41 publications do not report any), which suggests} strong governmental backing for the logistics sector that operates with low profit margins \parencite{zhang2023study, mckinsey2021lessons}. The motivation \hlone{for strong governmental backing} is the potential for technological advancements that will benefit national infrastructure and economic growth \parencite{xie2014china, owens2024china}. Besides this dominant public sector support, industry funding and public-industry collaborations constitute \hltwo{less than a tenth (6.32~\%)} of the publications, suggesting that while there is some industry involvement, companies largely rely on public sector investments to drive technological development \hlone{of \ac{ilu} identification systems. Nevertheless}, \hltwo{two studies (2.11~\%) are funded by industry alone, which may reflect an interest of companies} to invest in research to remain competitive \parencite{kwon2022logistics, parola2017drivers}.

\begin{figure}
    \centering
\begin{tikzpicture}
\begin{axis}[
  width=\linewidth, height=0.5\linewidth, font=\small,
  ybar stacked, bar width=7pt,
  symbolic x coords={1990,1999,2002,2003,2004,2006,2007,2009,2010,2011,2012,2013,2015,2016,2017,2018,2019,2020,2021,2022,2023,2024,2025,2026},
  xtick=data, x tick label style={rotate=45, anchor=east},
  enlarge x limits=0.03, ymin=0,
  xlabel={Year}, ylabel={Number of publications},
  ymajorgrids, grid style={black!15},
  legend style={at={(0.02,0.97)}, anchor=north west, legend columns=2, draw=black!30, font=\footnotesize}, legend cell align=left,
]
\addplot[fill=okabeOrange, draw=none] coordinates {(1990,0) (1999,0) (2002,0) (2003,1) (2004,0) (2006,0) (2007,0) (2009,0) (2010,0) (2011,0) (2012,0) (2013,0) (2015,0) (2016,0) (2017,0) (2018,0) (2019,0) (2020,0) (2021,0) (2022,0) (2023,0) (2024,2) (2025,1) (2026,0)};
\addlegendentry{A}
\addplot[fill=okabeSkyBlue, draw=none] coordinates {(1990,1) (1999,1) (2002,1) (2003,0) (2004,0) (2006,0) (2007,0) (2009,1) (2010,0) (2011,0) (2012,0) (2013,0) (2015,0) (2016,0) (2017,0) (2018,0) (2019,0) (2020,1) (2021,0) (2022,0) (2023,0) (2024,0) (2025,0) (2026,0)};
\addlegendentry{B}
\addplot[fill=okabeBluishGreen, draw=none] coordinates {(1990,0) (1999,0) (2002,0) (2003,0) (2004,0) (2006,0) (2007,0) (2009,0) (2010,0) (2011,0) (2012,0) (2013,0) (2015,0) (2016,1) (2017,0) (2018,0) (2019,0) (2020,0) (2021,0) (2022,0) (2023,0) (2024,0) (2025,1) (2026,0)};
\addlegendentry{C}
\addplot[fill=okabeYellow, draw=none] coordinates {(1990,0) (1999,0) (2002,0) (2003,0) (2004,0) (2006,0) (2007,0) (2009,0) (2010,0) (2011,2) (2012,0) (2013,0) (2015,0) (2016,0) (2017,0) (2018,0) (2019,0) (2020,1) (2021,2) (2022,1) (2023,1) (2024,3) (2025,6) (2026,1)};
\addlegendentry{Q1}
\addplot[fill=okabeBlue, draw=none] coordinates {(1990,0) (1999,0) (2002,0) (2003,0) (2004,0) (2006,1) (2007,0) (2009,0) (2010,0) (2011,0) (2012,0) (2013,0) (2015,0) (2016,1) (2017,0) (2018,0) (2019,0) (2020,0) (2021,0) (2022,0) (2023,1) (2024,1) (2025,3) (2026,0)};
\addlegendentry{Q2}
\addplot[fill=okabeVermillion, draw=none] coordinates {(1990,0) (1999,0) (2002,0) (2003,0) (2004,0) (2006,0) (2007,0) (2009,0) (2010,0) (2011,0) (2012,0) (2013,1) (2015,1) (2016,1) (2017,0) (2018,0) (2019,0) (2020,1) (2021,0) (2022,0) (2023,1) (2024,0) (2025,0) (2026,0)};
\addlegendentry{Q3}
\addplot[fill=okabeReddishPurple, draw=none] coordinates {(1990,0) (1999,0) (2002,0) (2003,0) (2004,1) (2006,0) (2007,0) (2009,0) (2010,0) (2011,0) (2012,1) (2013,1) (2015,0) (2016,0) (2017,0) (2018,0) (2019,0) (2020,0) (2021,0) (2022,0) (2023,2) (2024,0) (2025,0) (2026,1)};
\addlegendentry{Q4}
\addplot[fill=okabeBlack, draw=none] coordinates {(1990,0) (1999,0) (2002,0) (2003,2) (2004,0) (2006,2) (2007,2) (2009,0) (2010,1) (2011,2) (2012,0) (2013,1) (2015,2) (2016,2) (2017,3) (2018,4) (2019,6) (2020,5) (2021,2) (2022,1) (2023,3) (2024,4) (2025,4) (2026,3)};
\addlegendentry{Unranked}
\end{axis}
\end{tikzpicture}
    \caption{Ranking of publication outlets for the eligible articles (\hltwo{$n = 95$}) \hltwo{by publication year. A--C: CORE conference ranks; Q1--Q4: Scimago journal quartiles; Unranked: outlets listed in neither ranking.} The years with zero eligible publications are omitted in this bar chart. \hltwo{The data for 2026 are incomplete; they cover publications up to the search cut-off of 12 October 2026.}}
    \label{fig:ranks_by_year}
\end{figure}

While funding sources indicate industry and governmental priorities, the ranking of publication venues offers insight into the field’s academic recognition. To analyse the rankings of journals and conferences we followed the rankings of Scimago\footnote{\href{https://www.scimagojr.com/}{https://www.scimagojr.com/}} for journals and CORE\footnote{\href{https://www.core.edu.au/icore-portal}{https://www.core.edu.au/icore-portal}} for conferences. Figure~\ref{fig:ranks_by_year} shows that more than half of the articles on automatic \ac{ilu} identification \hltwo{(49 of 95)} are published in unranked outlets, which is not uncommon for applied or niche technical topics \parencite{chavarro2017researchers}. \hltwo{All 13 articles published between 2017 and 2019 appeared in unranked outlets. The share of articles in ranked outlets has since risen, from 23~\% (6 of 26) in 2016--2020 to 64~\% (25 of 39) in 2021--2025, as has the number of publications.} This suggests that the \ac{ilu} identification topic is relevant for high-quality research venues. Nevertheless, the field could benefit from a greater emphasis on more novel contributions, increased comparability, and reproducibility of the results. This could be achieved through \hlone{a common terminology,} publicly available benchmark datasets and repositories that includes the source code and the model weights. By adopting such research standards, \ac{ilu} identification tasks with \ac{cv}-based methods could be more relevant for high-ranked publication outlets.

\subsection{Terminology Beyond OCR-Gates}
\label{sec:terminology_beyon_ocr_gates}
Terminal entrances \hltwo{remain critical bottlenecks} in terms of operational efficiency and congestion impact\hlone{. Therefore,} their optimisation becomes one of the key challenges in terminal management. \hltwo{Most of} the earlier studies on this\hltwo{ applied} \ac{cv} methods at terminal gates, where the task is to automatically identify loading units as they \hltwo{enter or exit} the facility. In the literature, this task is \hlone{frequently} referred to as \textit{\aclu{accr}} (\textit{\acs{accr}}; \nptextcite{lau2024modelling, yu2024two, nguyen2020automatic, zhiming2019automatic, cao2017automatic, yoon2016automatic, verma2016automatic}) and rarely as \textit{loading unit identification} \parencite{teegen2024drone} or \textit{transportation unit identification} \parencite{gulsoylu2025trudi, gulsoylu2025inland}. \hltwo{In the titles of the 95 reviewed studies, 42 contain \textit{container code} and 18 \textit{container number}, so \textit{container number recognition} is a second established term. It shares the limitations of \ac{accr} discussed below.} \hltwo{The most common term, \ac{accr}, has two}\hlone{ limitations: (1) the word \textit{container} does not cover all types of loading units such as semi-trailers and swap bodies }\hltwo{that have ISO~6346-compliant}\hlone{ codes (as defined in EN 13044;} \nptextcite{UIRR2011new}\hlone{), (2) the word \textit{recognition} does not represent all }\hltwo{stages}\hlone{ involved (e.g., text detection)}. \hlone{Real-world scenarios, where text is typically embedded within complex and cluttered scenes, require an additional \textit{text detection} stage to localise the text regions before recognition can be performed} \parencite{Liu2023survey}. \hltwo{The term \textit{container identification}} \parencite{kumano2004development, liang2023performance}\hltwo{ has a further limitation: it is also used in several other domains, so a search for it returns a low share of eligible studies, and the term alone does not tell the reader which CV task is meant. Outside freight transport, it denotes, for example, the identification of specimen containers in clinical laboratories by bar codes} \parencite{clsi2005auto2}\hltwo{, of liquid containers by RFID tags} \parencite{kadono2021development}\hltwo{ and of copper canisters for spent nuclear fuel by ultrasonic measurements} \parencite{clementi2020identification}\hltwo{. Within freight transport, it often refers to identification by RFID tags} \parencite{petrariu2016analysis}\hltwo{ or to locating one container in a stack, such as the one in which a fire started} \parencite{konon2023image}\hltwo{, rather than to reading its ID code. In our search, Google Scholar estimated about 1,560 results for \textit{container identification}, more than five times as many as for any other of the search terms listed in Section~\ref{sec:methodology}, yet only 25 of its first 190 results (13~\%) were eligible studies, against 54 of the first 200 (27~\%) for \textit{container code recognition}.}

\begin{figure}[!h]
    \centering
    \includegraphics[width=0.9\linewidth]{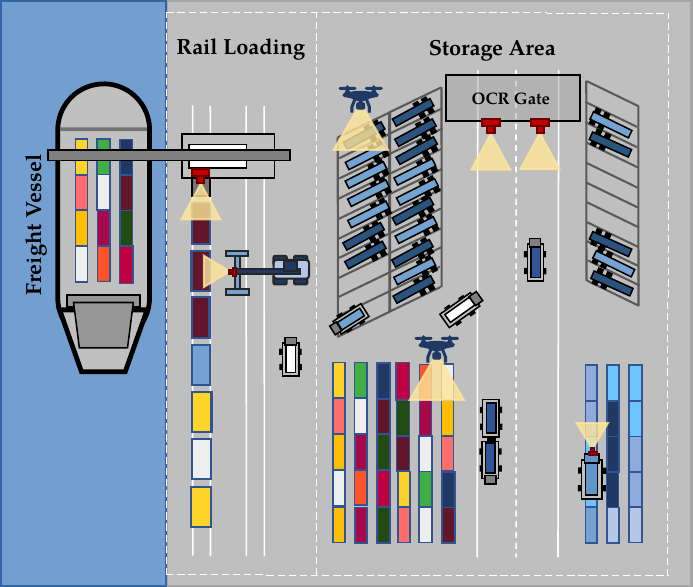}
    \caption{Schematic overview of a terminal with the \hltwo{operations in which \acp{ilu} are identified: vessel unloading at the quay crane, rail loading, storage} and \hltwo{yard transport, and gate operations. Fixed cameras at the \ac{ocr}-gate and on the crane are complemented by cameras on terminal} vehicles and \hltwo{drones, which extend identification from the gate to the whole terminal}.}
    \label{fig:port_figure}
\end{figure}

\hlone{Given the} \hlone{advancements in \ac{cv} and robotics}, \hlone{intermodal loading} \hltwo{unit} identification is now performed \hlone{beyond the gates,} \hlone{at quay cranes} \parencite{koo2012novel, wu2019key, cheng2025fast, choi2023design, kang2026vantage}\hlone{, terminal vehicles such as reach stackers} \parencite{gulsoylu2025trudi, gulsoylu2025inland} \hlone{and drones} \parencite{teegen2024drone, gulsoylu2025trudi, liu2025lightweight, gulsoylu2025inland} \hlone{as shown in Figure~\ref{fig:port_figure}. These systems can support different terminal operations, including vessel unloading, yard transportation, storage management, gate operations, and rail loading.} These new use cases introduce more variation and complexity than the original gate-based setting for containers. For these reasons, we propose the term \textit{Intermodal Loading Unit Identification} as an inclusive designation. This term covers various loading units and also \ac{ilu} detection/segmentation stage prior to text detection and recognition steps, introduced in \hltwo{four of the reviewed} studies \parencite{wu2019key, gulsoylu2025trudi, cheok2025maximizing}.

A term that covers all relevant \hltwo{stages}, aligned with \ac{cv} terminology\hltwo{,} is important, as it \hltwo{connects the task to} challenges shared across \ac{cv} applications. Using a common terminology can link the relevant studies to the broader problems. For example, detecting different scales of objects and text is becoming more relevant for \hltwo{the} \ac{ilu} identification task with the increasing use of mobile cameras \parencite{gulsoylu2025trudi}. However, small object segmentation \parencite{mun2024small} or small text detection \parencite{zheng2020scale} have been actively studied in \hltwo{general-purpose} \ac{cv} methods for many years. Moreover, a common terminology can help in finding the distinct aspects of \ac{ilu} identification from other \ac{cv} applications.

Unlike in common text recognition, where a language model based \hltwo{decoder} (\hltwo{e.g.,} TrOCR; \nptextcite{li2023trocr}) is useful due to contextual cues present in the natural language, the contextless nature of \hltwo{ISO~6346}-compliant ID codes \hltwo{removes this advantage. Reviewed studies report gains from language modules, but the gains are small (0.7--1.3 percentage points;} \nptextcite{cheng2025fast, cheok2025maximizing}\hltwo{) or measured on small test sets against baselines that differ in more than the language module (e.g., 81 test images;} \nptextcite{lau2024modelling}\hltwo{), and language-aware recognisers perform poorly on vertical codes} \parencite{zhu2025vertically, roumpattana2026enhancing}\hltwo{. Before fine-tuning, a recogniser pre-trained on natural language may even output words instead of ID codes} \parencite{teegen2024drone}\hltwo{, and such methods tend to over-rely on their language model} \parencite{garcia2024step}\hltwo{, which limits} their applicability in this context. Adopting the term \ac{ilu} identification can provide conceptual clarity and establish a coherent basis for researchers and practitioners. \hltwo{Supplementary Section~S2 defines the approaches and the \ac{cv} tasks that this review refers to.}

\subsection{Methods}\label{sec:methods}
\subsubsection{Classical Approaches}
The development of methods for \ac{ilu} identification has undergone a significant transformation over three decades, as illustrated in Figure~\ref{fig:publication_count_and_method_categories}. For two decades between 1990 and 2010, the field relied predominantly on \ac{dip} techniques combined with early \ac{ml} methods and\hltwo{,} to a lesser degree, \hltwo{on} \ac{dip} methods alone. This mirrors broader trends in the \ac{cv} literature \parencite{hsieh2024deep}. In this period, researchers mainly followed a three-stage approach: 1) Removal of the background using thresholding \parencite[e.g.,][]{gu2010efficient, zhiwei2003new, rosell2006segmentation, zini2009classification}, edge detection \parencite{zhiwei2003new, kim2006intelligent, kim2007fuzzy} and morphological operations \parencite{igual2002preprocessing, yin2007new, gu2010efficient}. 2) After segmenting the characters, each was classified by methods like template matching \parencite{kumano2004development}, or traditional \ac{ml} methods like \aclu{mlp} (\acs{mlp}; \nptextcite{rumelhart1986learning, lee1999automatic, koo2012novel})\hltwo{, nearest-neighbour search} \parencite{igual2002preprocessing}\hltwo{, the neocognitron} \parencite{lui1990neural}\hltwo{, ART2-based networks} \parencite{kim2006intelligent, kim2007recognition} \hltwo{or a fuzzy RBF network} \parencite{kim2007fuzzy}. 3) Finally, the predictions on the individual characters were post-processed to output a full ID code. These methods performed adequately in controlled image acquisition systems like \ac{ocr}-gates \parencite{Moszyk2021automation} but \hltwo{later studies report that they} struggled with challenging conditions, such as variability of brightness \parencite{nguyen2020automatic} and \hltwo{damage} \parencite{yao2025advanced, lin2024research} on the \ac{ilu} images.

\begin{figure}
    \centering
\begin{tikzpicture}
\begin{axis}[
  width=\linewidth, height=0.55\linewidth, font=\small,
  ybar stacked, bar width=18pt,
  symbolic x coords={1990--1995,1996--2000,2001--2005,2006--2010,2011--2015,2016--2020,2021--2025,2026},
  xtick=data, x tick label style={rotate=30, anchor=east},
  enlarge x limits=0.1, ymin=0, enlarge y limits={upper, value=0.1},
  xlabel={Time period}, ylabel={Number of publications},
  ymajorgrids, grid style={black!15},
  legend pos=north west, legend cell align=left, legend style={draw=black!30, font=\footnotesize},
]
\addplot[fill=okabeOrange, draw=none] coordinates {(1990--1995,1) (1996--2000,1) (2001--2005,3) (2006--2010,4) (2011--2015,8) (2016--2020,3) (2021--2025,2) (2026,0)};
\addlegendentry{DIP + ML}
\addplot[fill=okabeSkyBlue, draw=none] coordinates {(1990--1995,0) (1996--2000,0) (2001--2005,2) (2006--2010,3) (2011--2015,3) (2016--2020,3) (2021--2025,0) (2026,0)};
\addlegendentry{DIP}
\addplot[fill=okabeBluishGreen, draw=none] coordinates {(1990--1995,0) (1996--2000,0) (2001--2005,0) (2006--2010,0) (2011--2015,0) (2016--2020,7) (2021--2025,9) (2026,0)};
\addlegendentry{DIP + DL}
\addplot[fill=okabeYellow, draw=none] coordinates {(1990--1995,0) (1996--2000,0) (2001--2005,0) (2006--2010,0) (2011--2015,0) (2016--2020,11) (2021--2025,27) (2026,5)};
\addlegendentry{DL}
\addplot[fill=okabeVermillion, draw=none] coordinates {(1990--1995,0) (1996--2000,0) (2001--2005,0) (2006--2010,0) (2011--2015,0) (2016--2020,2) (2021--2025,1) (2026,0)};
\addlegendentry{DIP + ML + DL}
\node[anchor=south, font=\small] at (axis cs:1990--1995,1) {1};
\node[anchor=south, font=\small] at (axis cs:1996--2000,1) {1};
\node[anchor=south, font=\small] at (axis cs:2001--2005,5) {5};
\node[anchor=south, font=\small] at (axis cs:2006--2010,7) {7};
\node[anchor=south, font=\small] at (axis cs:2011--2015,11) {11};
\node[anchor=south, font=\small] at (axis cs:2016--2020,26) {26};
\node[anchor=south, font=\small] at (axis cs:2021--2025,39) {39};
\node[anchor=south, font=\small] at (axis cs:2026,5) {5};
\end{axis}
\end{tikzpicture}
    \caption{The number of publications and \hltwo{the method classes they use} for \hltwo{the} \ac{ilu} identification task over the years. \hltwo{The last period covers 2026 only, up to the search cut-off of 12 October 2026.}}
    \label{fig:publication_count_and_method_categories}
\end{figure}

In the subsequent five years, \hltwo{the number of} hybrid \ac{dip} and \ac{ml} \hltwo{studies doubled from four to eight} \parencite{fahn2015vision, wu2015new, chen2013container, sharma2013smart, koo2012novel, wu2012automated, chen2011computer, chen2011hidden}. This reflects the growing recognition of \ac{ml}’s potential for enhancing robustness in character and symbol recognition tasks in \ac{cv} \parencite{chauhan2024hcr}. However, the field lagged several years behind the broad adoption of \ac{dl}, despite its transformative influence on \ac{cv} solutions \parencite{murthy2020investigations, angelica2021impact} from 2012 onwards (e.g., the emergence of AlexNet; \nptextcite{krizhevsky2012imagenet}). This delay may be attributed to the scarcity of labelled \ac{ilu} identification datasets, limited computational infrastructure, lack of technology skills in the workforce, and the conservative approach for adopting new technologies in the logistics industry \parencite{evangelista2013technology, wohlleber2022prevailing}.

\subsubsection{Emergence of Deep Learning Approaches}
The five-year interval from 2016 to 2020 marks a turning point, with a \hltwo{notable adoption of} \ac{dl}-based methods. Publications using only \ac{dl}-based approaches rise sharply \hltwo{(11 of 26)}, as well as the hybrid strategies \hltwo{that combine \ac{dl} with \ac{dip} (seven) or with \ac{dip} and \ac{ml} (two)}. This phase was driven by advances in GPU hardware \parencite{jeon2021deep, pandey2022transformational}, and the availability of open-source frameworks such as TensorFlow \parencite{abadi2016tensorflow} and PyTorch \parencite{paszke2019pytorch}. Deep architectures, particularly \acused{cnn}\aclp{cnn} (\acsp{cnn}; \nptextcite{lecun1995convolutional}), \hltwo{were reported to outperform classical-method pipelines} for conducting \ac{ilu} identification under challenging conditions \parencite{verma2016automatic}. \hltwo{Meanwhile, the share of} purely \ac{dip}-based approaches declined\hltwo{ (3 of 11 studies in 2011--2015, 3 of 26 in 2016--2020, none since 2021), and \ac{dip} was} often limited to pre- or post-processing roles within hybrid pipelines \parencite{roeksukrungrueang2018implementation, liu2018container, zhiming2019automatic, zhang2020adaptive, tang2020container}.

\begin{table}
\centering
\newlength{\chipw}\settowidth{\chipw}{\scriptsize Recognition}
\newlength{\chipW}\setlength{\chipW}{\dimexpr\chipw+4pt\relax}
\newlength{\arrw}\setlength{\arrw}{0.7em}
\newlength{\iow}\settowidth{\iow}{\textbf{Output}}
\newcommand{\stageio}[1]{{\scriptsize\begin{tabular}{@{}c@{}}#1\end{tabular}}}
\newcommand{\stagechip}[3]{\tikz[baseline={([yshift=-0.5ex]c.center)}]{\node[#2, rounded corners=2pt, inner sep=2pt, text width=\dimexpr#1\chipW+#1\arrw-\arrw-4pt\relax, align=center, font=\scriptsize, execute at begin node={\hyphenpenalty=10000\relax}, minimum height=2.3em](c){#3};\path ([yshift=2.5pt]c.north) ([yshift=-2.5pt]c.south);}}
\newcommand{\stagearrow}[1]{\makebox[\dimexpr#1\chipW+#1\arrw+\arrw\relax]{\hspace{2pt}\rightarrowfill\hspace{2pt}}}
\renewcommand{\arraystretch}{1.25}
\caption{Stages defined in \ac{dl}-based pipelines for \ac{ilu} identification. The table reports how often each combination of stages appears among the \hltwo{62} \ac{dl}-based studies ($n = 62$). \hltwo{Each row is one pipeline with its stages in processing order, and each stage has a column and a colour of its own. Text spotting: one jointly trained network detects and recognises the text. Input: what the first stage takes (an image of the \ac{ilu} or of a part of it, a crop of the code region, or crops of single characters); output: what the last stage gives (boxes from detection, the ID code from recognition).}}
\label{tab:main_approach}
\begin{tabular}{@{}>{\centering\arraybackslash}p{\iow}@{}p{\arrw}@{}>{\centering\arraybackslash}p{\chipW}@{}p{\arrw}@{}>{\centering\arraybackslash}p{\chipW}@{}p{\arrw}@{}>{\centering\arraybackslash}p{\chipW}@{}p{\arrw}@{}>{\centering\arraybackslash}p{\chipW}@{}p{\arrw}@{}>{\centering\arraybackslash}p{\chipW}@{}p{\arrw}@{}>{\centering\arraybackslash}p{\iow}@{\hspace{0.4em}}r@{\,}r@{}}
\toprule
\textbf{Input} & & \multicolumn{9}{@{}c@{}}{\textbf{Pipeline (stages in processing order)}} & & \textbf{Output} & \multicolumn{2}{r@{}}{\begin{tabular}[b]{@{}r@{}}\textbf{Studies}\\\textbf{(\%)}\end{tabular}} \\
\midrule
\stageio{Image} & \multicolumn{3}{@{}l@{}}{\stagearrow{1}} & \stagechip{1}{fill=okabeSkyBlue!60}{Text\strut\\detection\strut} & \stagearrow{0} & \stagechip{1}{fill=okabeBluishGreen!60}{Text\strut\\recognition\strut} & \multicolumn{5}{@{}l@{}}{\stagearrow{2}} & \stageio{ID code} & 27 & (43.5) \\  %
\stageio{Image} & \multicolumn{3}{@{}l@{}}{\stagearrow{1}} & \stagechip{1}{fill=okabeSkyBlue!60}{Text\strut\\detection\strut} & \multicolumn{3}{@{}l@{}}{\stagearrow{1}} & \stagechip{1}{fill=okabeYellow!60}{Character\strut\\detection\strut} & \stagearrow{0} & \stagechip{1}{fill=okabeBlue!60}{Character\strut\\recognition\strut} & \stagearrow{0} & \stageio{ID code} &  8 & (12.9) \\  %
\stageio{Image} & \multicolumn{3}{@{}l@{}}{\stagearrow{1}} & \stagechip{1}{fill=okabeSkyBlue!60}{Text\strut\\detection\strut} & \multicolumn{7}{@{}l@{}}{\stagearrow{3}} & \stageio{Code box} &  6 & (9.7) \\  %
\stageio{Image} & \multicolumn{3}{@{}l@{}}{\stagearrow{1}} & \multicolumn{3}{@{}c@{}}{\stagechip{2}{left color=okabeSkyBlue!60, right color=okabeBluishGreen!60}{Text spotting}} & \multicolumn{5}{@{}l@{}}{\stagearrow{2}} & \stageio{ID code} &  5 & (8.1) \\  %
\stageio{Image} & \multicolumn{7}{@{}l@{}}{\stagearrow{3}} & \stagechip{1}{fill=okabeYellow!60}{Character\strut\\detection\strut} & \stagearrow{0} & \stagechip{1}{fill=okabeBlue!60}{Character\strut\\recognition\strut} & \stagearrow{0} & \stageio{ID code} &  5 & (8.1) \\  %
\stageio{Image} & \multicolumn{3}{@{}l@{}}{\stagearrow{1}} & \stagechip{1}{fill=okabeSkyBlue!60}{Text\strut\\detection\strut} & \multicolumn{5}{@{}l@{}}{\stagearrow{2}} & \stagechip{1}{fill=okabeBlue!60}{Character\strut\\recognition\strut} & \stagearrow{0} & \stageio{ID code} &  5 & (8.1) \\  %
\stageio{Image} & \stagearrow{0} & \stagechip{1}{fill=okabeOrange!60}{Object\strut\\detection\strut} & \stagearrow{0} & \stagechip{1}{fill=okabeSkyBlue!60}{Text\strut\\detection\strut} & \stagearrow{0} & \stagechip{1}{fill=okabeBluishGreen!60}{Text\strut\\recognition\strut} & \multicolumn{5}{@{}l@{}}{\stagearrow{2}} & \stageio{ID code} &  3 & (4.8) \\  %
\stageio{Character\\crops} & \multicolumn{9}{@{}l@{}}{\stagearrow{4}} & \stagechip{1}{fill=okabeBlue!60}{Character\strut\\recognition\strut} & \stagearrow{0} & \stageio{ID code} &  1 & (1.6) \\  %
\stageio{Image} & \multicolumn{7}{@{}l@{}}{\stagearrow{3}} & \stagechip{1}{fill=okabeYellow!60}{Character\strut\\detection\strut} & \multicolumn{3}{@{}l@{}}{\stagearrow{1}} & \stageio{Character\\boxes} &  1 & (1.6) \\  %
\stageio{Code crop} & \multicolumn{5}{@{}l@{}}{\stagearrow{2}} & \stagechip{1}{fill=okabeBluishGreen!60}{Text\strut\\recognition\strut} & \multicolumn{5}{@{}l@{}}{\stagearrow{2}} & \stageio{ID code} &  1 & (1.6) \\  %
\bottomrule
\end{tabular}
\end{table}

Until 2016, all methods developed for \ac{ilu} identification focused on character-level localisation and recognition\hltwo{, and every study of that period that describes its camera used a fixed one}. This approach is still in use with \ac{dl} methods\hltwo{, in 20 studies, 12 of them since 2021} \parencite[e.g.,][]{diaz2024wagon, ngo2023vision, liu2025faster, samsudin2026freight, roumpattana2026enhancing}\hltwo{,} as \ac{ilu} identification on fixed cameras (Figure~\ref{fig:perspectives}\hltwo{, left}) often resembles \ac{ocr} on scanned documents. However, with the increasing use of vehicle-mounted cameras (Figure~\ref{fig:perspectives}\hltwo{, middle and right}), the task has evolved to resemble scene-text detection and recognition rather than \ac{ocr}. \hltwo{At the same time}, the trend is shifting towards using comprehensive scene-text detection and recognition methods rather than relying on character detection and recognition. Table~\ref{tab:main_approach} shows that the majority of articles employ \ac{dl} methods for both scene-text detection and recognition tasks. \hltwo{Still, 19 of the 62 studies that use \ac{dl} combine it with classical image processing, three of them also with machine learning (Figure~\ref{fig:publication_count_and_method_categories}), for a step of the pipeline or for check-digit verification, which is coded as \ac{dip} (Supplementary Section~S1)}. Examples include using traditional techniques for text detection such as cropping a predefined area for searching ID codes \parencite{hoa2023build} or for character recognition, template matching after a \ac{dl} model detects text areas \parencite{mei2016novel}. \hltwo{Most studies (43 of 62), however,} favour more comprehensive pipelines that rely on \ac{dl} in each step.

\subsubsection{Deep Learning Domination}
By 2021–2025, \ac{dl}-based approaches had established themselves as the predominant paradigm for \ac{ilu} identification. \hltwo{All five studies from 2026, up to the search cut-off, use deep learning only. Transformers arrived late: 13 of the 14 studies that use them appeared in 2024--2026. Most \ac{dl} pipelines remain multi-stage, with two or more separately run stages (44 of 62 studies).} \hltwo{All five text-spotting models appeared in 2021 or later.} \hltwo{The share of studies that use \ac{dip} fell from 15 of 26 in 2016--2020 to 12 of 39 in 2021--2025. 10 of these 12 combine it with \ac{dl}, and only two use no \ac{dl} at all. Studies of this period argue that \ac{dl} models generalise better than handcrafted pipelines under varying conditions} \parencite{lau2024modelling, yang2023lightweight, zhang2021vertical}\hltwo{. Higher accuracy, however, is not visible in the reported results due to the lack of a common benchmark dataset. The median end-to-end accuracies of the method classes are comparable (Section~\ref{sec:evaluation}). Classical methods remain in use, alone or as steps of \ac{dl} pipelines,} in specialised scenarios such as real-time \hltwo{or resource} constraints \parencite{ademola2023resource, li2022towards, geirinhas2025divide}. Researchers also address real-time constraints by focusing on text spotting \parencite{zhao2024practical, chen2026end}\hltwo{,} as these methods are generally faster and more efficient due to their unified feature extraction module \parencite{long2022towards, jaderberg2014deep, zhang2022text}. \hltwo{Four} studies introduce object detection or instance segmentation as a prior step to text detection and recognition\hltwo{, three of them with \ac{dl}} \parencite{gulsoylu2025trudi, gulsoylu2025inland, cheok2025maximizing} \hltwo{and one with classical methods} \parencite{wu2019key}. Detecting the \ac{ilu} reduces the search area for the text detection step and enables association of detected and recognised IDs with \hltwo{the detected \acp{ilu}}, thereby improving scene understanding \parencite{gulsoylu2025trudi}.

\subsection{Datasets}\label{sec:datasets}
The quality and diversity of annotated data strongly influence how well \ac{dl} models perform once deployed in real-world port environments. Carefully curated datasets with consistent labelling not only enable robust training and evaluation, but also determine whether methods can generalise beyond laboratory settings. Understanding the properties of datasets is therefore essential for evaluating the performance and applicability of \ac{cv}-based \ac{ilu} identification systems. Below, we summarise the main characteristics of the datasets used in the eligible studies, highlighting camera position settings, diversity, and availability.

\subsubsection{Image Acquisition Settings}
\begin{figure}
    \centering
    \includegraphics[height=3.75cm]{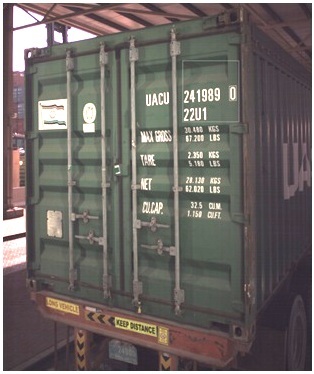}
    \includegraphics[height=3.75cm]{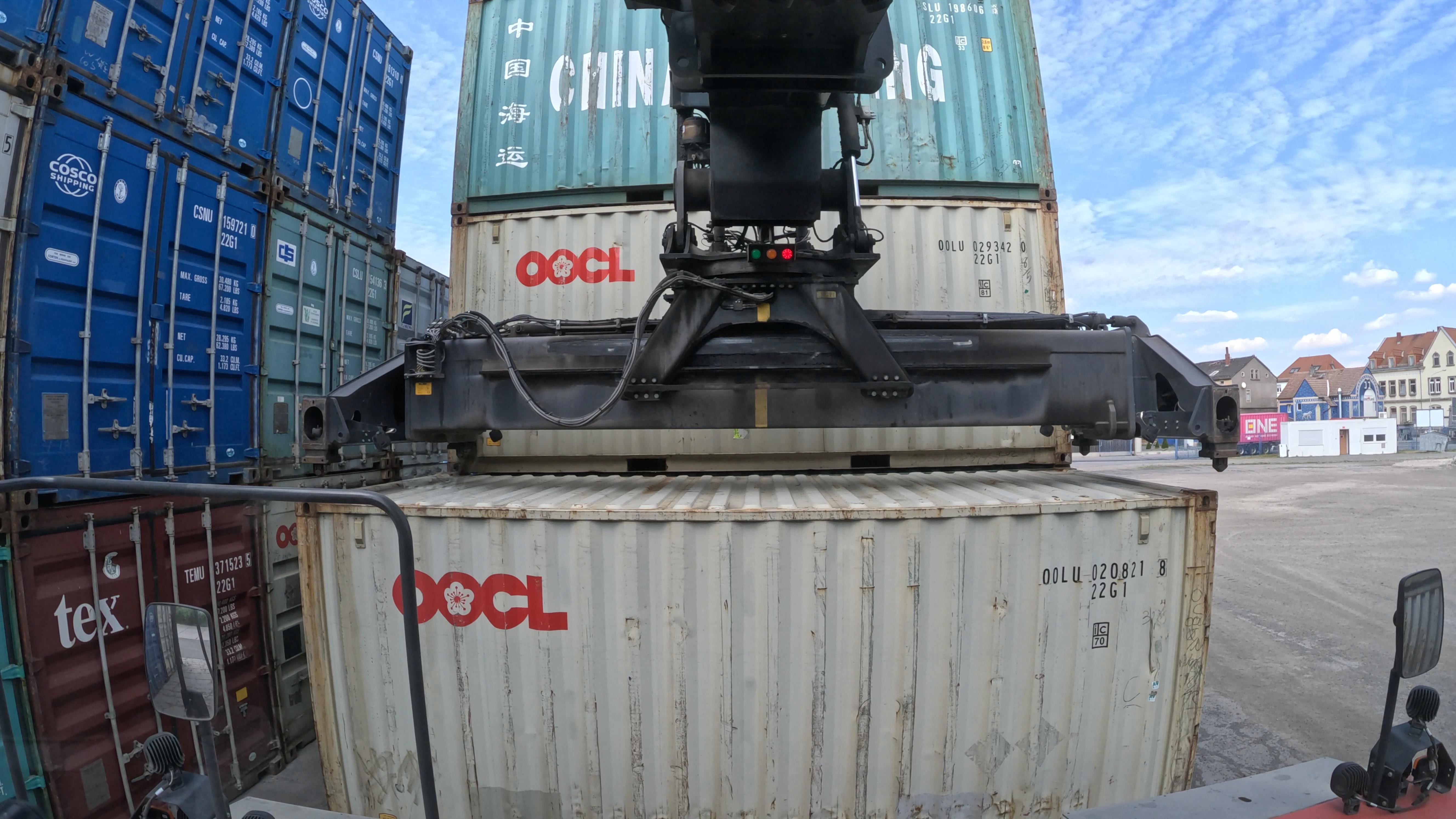}
    \includegraphics[height=3.75cm]{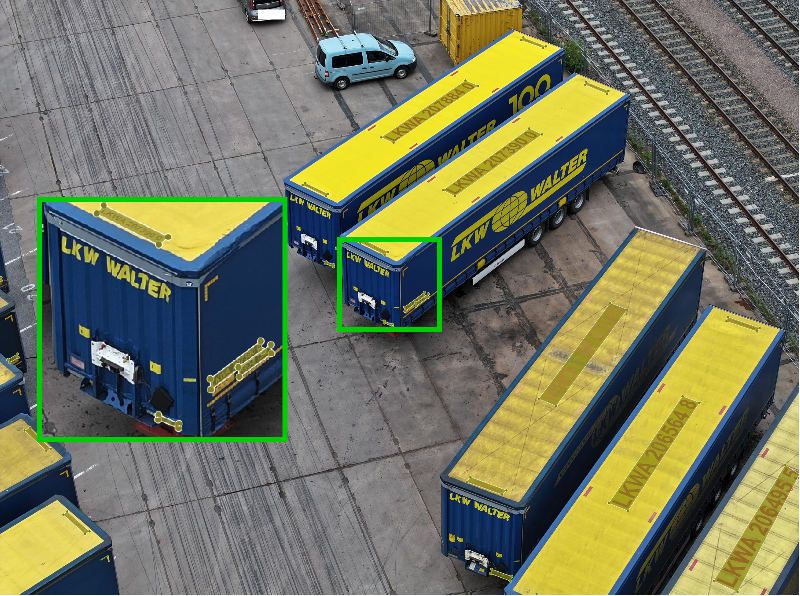}
    \caption{Sample images for fixed perspective (left; \nptextcite{verma2016automatic}), ground vehicle perspective (middle; \nptextcite{gulsoylu2025trudi}) \hltwo{and} aerial perspective (right; \nptextcite{teegen2024drone}).}
    \label{fig:perspectives}
\end{figure}

We categorised cameras as \textit{fixed} or \textit{mobile cameras} based on their mounting and usage \hltwo{(}Figure~\ref{fig:perspectives}\hltwo{)}. Fixed cameras include \hltwo{\ac{ocr}-gates}, CCTV cameras, and crane-mounted cameras. These are installed at permanent locations such as terminal gates, yards, or crane structures, providing stable, controlled views for capturing \ac{ilu} codes. Their main advantage is delivering consistent, high-quality images ideal for computationally efficient methods like classical \ac{dip} and traditional \ac{ml}, as long as \acp{ilu} are present at predefined areas. \hltwo{Besides requiring substantial infrastructure investment for their installation and operation,} fixed cameras often struggle to capture sufficient feature information in large and complex environments, resulting in reduced accuracy and reliability of detection and recognition based on visual neural networks \parencite{shen2024identification}.

In contrast, mobile cameras are mounted on \acp{uav}, reach stackers, terminal trucks, or other industrial vehicles and smart devices. For ground-level perspectives, we also considered handheld cameras, as \hltwo{their images} closely resemble those captured by vehicle-mounted systems. \hltwo{Accordingly,} Table~\ref{tab:summary_camera_characteristics} \hltwo{splits mobile cameras into ground cameras (vehicle-mounted or handheld) and aerial cameras (mounted on \acp{uav}).} \hlone{The challenges posed by mobile settings, in contrast to static camera systems, include varying camera distances, changing viewpoints, occlusions and }\hltwo{real-time}\hlone{ processing constraints in congested terminal environments.} This mobility introduces higher image variability, requiring advanced processing methods and diverse datasets. \hltwo{When these challenges are addressed}, \hlone{there is a potential for more precise and real-time terminal monitoring. Thanks to }\hltwo{sensor-equipped}\hlone{ vehicles} \parencite{munuzuri2020using, cuong2024safe}\hlone{, \ac{cv} pipelines beyond \ac{ilu} identification can be created, such as localisation of \acp{ilu} through pose estimation, autonomous path optimisation for vehicles navigating around terminals}\hltwo{ and}\hlone{ multi-agent coordination (collaboration of aerial and ground vehicles) for terminal monitoring}.

\hlone{Critically, the feasibility of vehicle-mounted setups does not require the full \ac{ilu} identification pipeline to run on the vehicle itself. Two architectures are viable: a streaming setup in which frames are forwarded to a central server for inference on standard infrastructure, and an edge-to-cloud offloading setup in which a lightweight on-vehicle model selects legible frames and forwards only those to the server. This decouples recognition accuracy from the on-board compute envelope.}

\begin{table}
\centering
\caption{Summary of dataset characteristics: camera position, \ac{ilu} types covered\hltwo{, origin of the images} and dataset availability.}
\label{tab:summary_camera_characteristics}
\begin{tabular}{>{\centering\arraybackslash}m{1cm} lrr}
\toprule
\textbf{} & \textbf{Category} & \textbf{Count} & \textbf{Percentage} \\
\midrule
\multirow{7}{*}{\rotatebox[origin=c]{90}{Camera position}}
& Only Fixed & 47 & 49.47~\% \\
& Not reported & 39 & 41.05~\% \\
& Fixed and Ground & 3 & 3.16~\% \\
& Ground and Aerial & 2 & 2.11~\% \\
& Only Ground & 2 & 2.11~\% \\
& Fixed, Ground and Aerial & 1 & 1.05~\% \\
& Only Aerial & 1 & 1.05~\% \\
\midrule
\multirow{6}{*}{\rotatebox[origin=c]{90}{ILU types}}
& Containers only & 88 & 92.63~\% \\
& Containers, tank containers and semi-trailers & 2 & 2.11~\% \\
& Containers and semi-trailers & 2 & 2.11~\% \\
& Containers and railway wagons & 2 & 2.11~\% \\
& Containers and tank containers & 1 & 1.05~\% \\
& Including swap bodies & 0 & 0.00~\% \\
\midrule
\multirow{4}{*}{\rotatebox[origin=c]{90}{Origin}}
& Own & 71 & 74.74~\% \\
& Not reported & 15 & 15.79~\% \\
& Third-party & 5 & 5.26~\% \\
& Own and third-party & 4 & 4.21~\% \\
\midrule
\multirow{5}{*}{\rotatebox[origin=c]{90}{Availability}}
& Not public & 78 & 82.11~\% \\
& Upon request & 6 & 6.32~\% \\
& Fully available & 5 & 5.26~\% \\
& Partially available & 4 & 4.21~\% \\
& Behind paywall & 2 & 2.11~\% \\
\bottomrule
\end{tabular}
\end{table}

As shown in Table~\ref{tab:summary_camera_characteristics}\hltwo{, the most common collection setting (49.47~\%) is fixed cameras only, such as} CCTV cameras at \ac{ocr}-gates \parencite[e.g.,][]{zhao2024practical, ta2025advanced, diaz2024wagon, mi2020port} \hltwo{or on} cranes \parencite{wu2019key, koo2012novel, cheng2025fast, choi2023design, kang2026vantage}\hltwo{. However, 39 studies (41.05~\%) do not mention how their images were acquired: the camera position was coded only when the study describes the set-up or shows it unambiguously in a figure of its own data. A mixture of fixed and ground-level cameras is rare (3.16~\%;} \nptextcite{yu2024two, xu2024data, chen2026end}\hltwo{).} \hltwo{Meanwhile,} vehicle-mounted cameras are increasingly used alongside fixed cameras to capture wider areas and provide information across the terminal \parencite{teegen2024drone, liu2025lightweight, gulsoylu2025trudi, gulsoylu2025inland, chen2026end}.

\subsubsection{Diversity}
Most datasets consist of images gathered from \hlone{ports, reflecting their operational} settings. \hlone{Images in these datasets were taken from} the world's busiest ports like Shanghai \parencite{zhao2024practical, zhang2024enhancing}, \hltwo{Ningbo-Zhoushan} \parencite{cheng2025fast}\hltwo{,} Singapore \parencite{lui1990neural, hoa2023build}, Hong Kong \parencite{lee1999automatic}, Kaohsiung \parencite{tai2011method}, and Hamburg \parencite{gulsoylu2025trudi}, as well as \hltwo{Nanjing} \parencite{li2022towards}\hltwo{, Tianjin} \parencite{lin2024research}\hltwo{, Taicang} \parencite{mi2020port} and other ports in \hltwo{China (31 studies in all, including the mainland ports above)} \parencite[e.g.,][]{hoa2023build, liang2020research, wu2019key, wan2021multichannel}\hltwo{,} Malaysia \parencite{lau2024modelling, cheok2025maximizing, samsudin2026freight}, Sweden \parencite{diaz2024wagon}, Vietnam \parencite{ngo2023vision, nguyen2023digital, nguyen2020automatic, ta2025advanced}, Thailand \parencite{roeksukrungrueang2018implementation, roumpattana2026enhancing}, South Korea \parencite{choi2023design, kang2026vantage, yoon2016automatic, sharma2013novel, sharma2013smart, koo2012novel}, Japan \parencite{kumano2004development}, \hltwo{Indonesia} \parencite{safitri2026vgg16}\hltwo{, Portugal} \parencite{santos2025automating}\hltwo{, Spain} \parencite{rosell2006segmentation, igual2002preprocessing, garcia2024step}\hltwo{,} Italy \parencite{goccia2003recognition, zini2009classification}\hltwo{, Taiwan} \parencite{liu2025faster}\hltwo{, Canada} \parencite{zhang2020adaptive, zhang2021vertical} \hltwo{and Estonia} \parencite{ademola2023resource}. \hlone{While rare, there are datasets with images collected in} smaller inland ports like Dresden and Riesa \parencite{teegen2024drone, gulsoylu2025trudi, gulsoylu2025inland}. Besides the terminal-specific data collection, web scraping was also used \parencite{yang2023lightweight, verma2016automatic, wu2026container, han2018application, zhu2025vertically}. However, datasets consisting of images from multiple locations are rare \parencite[e.g.,][]{hoa2023build, ta2025advanced, gulsoylu2025trudi}. For \ac{dl} methods, capturing the nuances of \acp{ilu} across different countries and their respective ports is important.

The number of images used in the experiments in the eligible articles ranges from a minimum of 30 images to a maximum of \hltwo{34,000}, with a median of \hltwo{1,292} images \hltwo{(82 studies, counting full, original \ac{ilu} images only, without crops or augmented copies). The test sets are much smaller: in the 71 studies that state their test set in images, the median size is 500 images. The full datasets have a median of about 300 images in the 22 of these 82 studies published before 2016, none of which uses deep learning, and of 3,000 images in the 53 studies that use deep learning; the other seven are later studies without deep learning.} Since the \ac{dip}-based methods do not require a large amount of data, researchers who tackled \hltwo{the} \ac{ilu} identification task before \ac{dl} experimented with a lower number of images compared to recent studies. However, with the \hltwo{adoption of} \ac{dl}, the need for data has increased significantly. \hltwo{Several} studies have addressed this need by pre-training their models on general-purpose scene-text detection and recognition datasets such as ICDAR2015 \parencite{karatzas2015icdar}, CTW1500 \parencite{yuliang2017detecting}, and SynthText \parencite{gupta2016synthetic}, then fine-tuning on smaller \ac{ilu} datasets \parencite{lau2024modelling, teegen2024drone, gulsoylu2025trudi, zhang2019deep, li2019automatic, zhu2025vertically}.

Starting from the early datasets, efforts to represent diverse conditions have been made in order to evaluate generalisation. Lighting variability is \hltwo{one of} the most discussed \hltwo{factors (74 of the 95 studies report a lighting condition)}, with datasets often including images taken in bright, low-light, and uneven lighting scenarios \parencite[e.g.,][]{he2005new, zhiwei2003new}. Considering how the earlier, thresholding-based methods get affected by small changes in light intensity, diversifying the lighting conditions in the experiments was a major focus\hltwo{ (12 of the 14 studies published before 2011)} \parencite[e.g.,][]{goccia2003recognition, igual2002preprocessing, lee1999automatic}. However, currently lighting conditions are classified with more general classes: Day and night settings are frequently included to test the adaptability of the systems\hltwo{ (28 studies)} \parencite[e.g.,][]{yao2025advanced, liu2025lightweight, gulsoylu2025trudi, zhao2024practical}.

Weather conditions such as rain, snow, and fog are also represented, as they can significantly affect image quality and \ac{ilu} identification performance\hltwo{ (30 of the 95 studies report a weather condition. 17 of them name rain, snow or fog, the others only changing sky conditions or unspecified weather)} \parencite[e.g.,][]{lee1999automatic, zhao2021improved, zhang2021vertical, yang2023lightweight, gulsoylu2025trudi}. Additionally, some datasets feature images with motion blur, camera noise, dirty, rusty, damaged, or partially occluded ID codes to further challenge algorithms with the conditions that exist in real-world settings\hltwo{ (79 studies report at least one of them, but only 12 name motion blur and 26 occlusion;} Section~\ref{sec:failure_modes}\hltwo{)} \parencite[e.g.,][]{yao2025advanced, gulsoylu2025trudi, lin2024research, zhang2021vertical, verma2016automatic}. Rarely were these conditions simulated to increase the diversity of datasets \parencite{yang2023lightweight, zhao2021improved, lin2024research}.

\hltwo{One important limitation of the datasets is their coverage of \ac{ilu} types. All 95 studies work with container images. Only four of them (4.2~\%) also include semi-trailers} \parencite{zhang2021vertical, teegen2024drone, gulsoylu2025trudi, gulsoylu2025inland}\hltwo{, three (3.2~\%) contain tank containers} \parencite{gulsoylu2025trudi, gulsoylu2025inland, yu2024two}\hltwo{, and none treats swap bodies as a class of its own. Two studies identify railway wagon numbers next to container codes} \parencite{garcia2024step, diaz2024wagon}\hltwo{. Even the most class-rich public dataset,} \aclu{trudi} (\acs{trudi}, \nptextcite{gulsoylu2025trudi})\hltwo{, has no swap-body class and annotates swap bodies as containers. Moreover, no study reports class-wise performance for \ac{ilu} types. The only class-wise figures separate wagon numbers from container codes} \parencite{garcia2024step, diaz2024wagon}.

\subsubsection{Availability}

We divided datasets into 5 different categories: Fully available \parencite{gulsoylu2025trudi, liang2025dimension, geirinhas2025divide, garcia2024step, hlabisa2024automated}, upon request \parencite{chen2026end, liu2025lightweight, yao2025advanced, zhu2025vertically, xu2024data, yang2023lightweight}, partially available \parencite{khuboni2025confidence, santos2025automating, lin2024research, verma2016automatic}, behind paywall \parencite{kang2026vantage, choi2023design} and not public. Table~\ref{tab:summary_camera_characteristics} shows the availability \hltwo{per study}. We were not able to confirm the availability of the datasets that claim to be available upon request as our requests went unanswered.

\hltwo{The images of 78 of the 95 studies (82.11~\%) are not public}, which creates a substantial barrier for research and development in this area. Recent papers acknowledge the lack of benchmark datasets and its negative effects on the field \parencite{liu2025lightweight, gulsoylu2025trudi, yao2025advanced, xu2024data, khuboni2025confidence, liang2023performance}. The lack of availability prevents the reproducibility of results and fair comparison between the proposed methods for researchers. As for the \hlone{practitioners, while large seaports }\hltwo{may have}\hlone{ extensive} \hltwo{proprietary datasets}, publicly available datasets \hltwo{could extend} these capabilities to smaller terminals, facilitating data-driven optimisation.

\sloppy{
\hlone{After screening the literature, we could find} \hltwo{only eight publicly available datasets: four released by reviewed studies and four third-party datasets that reviewed studies use (Table~\ref{tab:public_datasets})}. Among \hltwo{the released ones}, the \textit{\ac{trudi}} dataset \parencite{gulsoylu2025trudi} stands out for its annotation density. It offers 35,034 instances across 733 images, covering five classes: \texttt{Container}, \texttt{trailer}, \texttt{tank container}, \texttt{ID code}, and \texttt{logo}. It introduces motion blur as an additional challenging condition, as most of its images were captured via \acp{uav}, reach stackers, and terminal trucks. \hltwo{The other three are a subset of the ContainerText dataset} \parencite{xu2024data}\hltwo{, released with the DVLT model} \parencite{liang2025dimension}\hltwo{, the STEP benchmark, a test set that is released in full and also contains railway wagon numbers} \parencite{garcia2024step}\hltwo{, and the test set of a container dataset with images from the web} \parencite{verma2016automatic}\hltwo{. The four third-party datasets are published on Roboflow Universe, GitHub and a shared drive. They are used, alone or together with the authors' own images, by} \textcite{geirinhas2025divide}\hltwo{,} \textcite{hlabisa2024automated}\hltwo{,} \textcite{khuboni2025confidence}\hltwo{,} \textcite{lin2024research} \hltwo{and} \textcite{santos2025automating}\hltwo{. These eight datasets account for the nine studies whose images are fully or partially available} \hltwo{(}Table~\ref{tab:summary_camera_characteristics}\hltwo{): four release all or part of their own images and five use the third-party datasets.}
}

\begin{table}
\centering
\caption{Publicly available \ac{ilu} identification datasets. \hltwo{Images and ID codes, conditions and location are given for the released datasets. The third-party datasets are listed with the studies that use them.}}
\label{tab:public_datasets}
\footnotesize
\setlength{\tabcolsep}{3pt}
\begin{tabularx}{\textwidth}{@{}>{\raggedright\arraybackslash}X
                r
                r
                >{\raggedright\arraybackslash}p{2.5cm}
                >{\raggedright\arraybackslash}p{3.1cm}
                >{\raggedright\arraybackslash}p{2.05cm}@{}}
\toprule
\textbf{Dataset} & \textbf{Images} & \textbf{ID codes} & \textbf{ILU classes} & \textbf{Conditions} & \textbf{Location} \\
\midrule
\multicolumn{6}{@{}l}{\textit{Released by a reviewed study}} \\
\midrule
\href{https://github.com/yikuizhai/DVLT}{ContainerText (DVLT subset)}\newline \textcite{liang2025dimension}
  & 1,600 & \begin{tabular}[t]{@{}r@{}}Not\\ reported\end{tabular} & Containers
  & Day/night\newline Vertical codes
  & Not reported \\
\midrule
\href{https://github.com/egulsoylu/trudi}{TRUDI}\newline \textcite{gulsoylu2025trudi}
  & 733 & \begin{tabular}[t]{@{}r@{}}Not\\ reported\end{tabular} & Containers,\newline tank containers,\newline semi-trailers
  & Day/night\newline Rainy, snowy\newline Motion blur\newline Damaged ID\newline Various angles, vertical codes
  & Several countries \\
\midrule
\href{https://github.com/CVC-DAG/STEP}{STEP test set}\newline \textcite{garcia2024step}
  & \begin{tabular}[t]{@{}r@{}}Not\\ reported\end{tabular} & \begin{tabular}[t]{@{}r@{}}Not\\ reported\end{tabular} & Containers,\newline railway wagons
  & Not reported
  & Spain \\
\midrule
\href{https://github.com/ankitjnu/TCS_Ship_Container_Dataset}{TCS Ship Container Dataset}\newline \textcite{verma2016automatic}
  & \begin{tabular}[t]{@{}r@{}}Not\\ reported\end{tabular} & \begin{tabular}[t]{@{}r@{}}Not\\ reported\end{tabular} & Containers
  & Varying light\newline Rainy, snowy, foggy\newline Dirty, rusty, faded ID\newline Various angles, vertical codes
  & Not reported \\
\midrule
\multicolumn{6}{@{}l}{\textit{Third-party datasets used by reviewed studies}} \\
\midrule
\href{https://universe.roboflow.com/dissertation-t3buk/container-number-detection}{Container Number Detection (Roboflow)} & \multicolumn{5}{@{}l@{}}{used by \textcite{geirinhas2025divide}} \\
\midrule
\href{https://github.com/lbf4616/ContainerNumber-OCR}{Container Number-OCR} & \multicolumn{5}{@{}l@{}}{used by \textcite{lin2024research}, \textcite{santos2025automating}} \\
\midrule
\href{https://universe.roboflow.com/project-ha37k/new-container-number}{New-container-number (Roboflow)} & \multicolumn{5}{@{}l@{}}{used by \textcite{hlabisa2024automated}, \textcite{khuboni2025confidence}} \\
\midrule
\href{https://drive.google.com/drive/folders/13LpHEeFExmDJnw_U9peqLR-8uAAUMEzi}{Shipping Container Code} & \multicolumn{5}{@{}l@{}}{used by \textcite{khuboni2025confidence}} \\
\bottomrule
\end{tabularx}
\end{table}

\subsection{Evaluation}\label{sec:evaluation}
To evaluate the proposed \ac{ilu} identification methods, several evaluation metrics are commonly used. As there are multiple stages of \ac{ilu} identification using \ac{cv} methods, such as text detection and text recognition, each stage can be evaluated individually. In the literature, \hltwo{69 of 95 articles (72.63~\%) report no end-to-end accuracy. Here, end-to-end accuracy is the printed rate of complete ID codes read correctly by the whole pipeline on the whole test set, whether the pipeline consists of several stages or of a single text-spotting model (Supplementary Section~S1). These 69 studies report the results of individual stages only, or end-to-end figures that do not meet this definition.} However, the results for individual stages offer limited insight \hltwo{into} the applicability of the method for the \ac{ilu} identification task. End-to-end accuracy or success rate is the most relevant metric due to its practical importance. It is also the most strict metric as one unrecognised character is enough for failed identification.

Precision and recall are the common metrics for these individual stages. These metrics show the balance between false positives and false negatives, with precision focusing on the accuracy of positive predictions and recall on the model's ability to identify all relevant instances. The F1-score, or H-mean, \hltwo{combines} precision and recall into a single metric, offering a balanced view of the model's performance.

For object detection, instance segmentation or text detection tasks, \ac{ap} and \ac{map} are commonly used, measuring precision across various \ac{iou} levels. The \ac{iou} metric is important for assessing the overlap between predicted and ground truth bounding boxes. As for the recognition, besides precision, recall and F1-score, other metrics such as the \ac{cer} \hlone{or word accuracy are} used. \ac{cer} measures the proportion of characters that were incorrectly predicted compared to the total number of characters in the reference text, using the \hltwo{Levenshtein distance} \parencite{levenshtcin1966binary}.

Speed and efficiency are evaluated using metrics such as \ac{fps} and inference time, which assess the model's operational speed. Model complexity is assessed through the number of parameters, \acp{gflop}, and model size, indicating the computational demands and storage requirements. \hlone{Similar to the accuracy metrics, the field's reporting practices around computational efficiency are uneven. Across the 95 eligible articles, inference speed is reported in }\hltwo{60 studies (63.2~\%), but a frame rate of the whole pipeline only in 27 (28.4~\%)}\hlone{, and at least one hardware specification (CPU, GPU, or RAM) is reported in }\hltwo{62 studies (65.3~\%)}\hlone{. Model size (parameter count or storage footprint) appears in only }\hltwo{16 studies (16.8~\%)}\hlone{ and computational complexity (FLOPs/GFLOPs) in only }\hltwo{9 (9.5~\%)}\hlone{. The reported hardware spans }\hltwo{more than two decades of platforms from a Pentium II at 333 MHz} \parencite{kumano2004development} \hlone{to consumer GPUs such as RTX 3090} \parencite{lau2024modelling}\hlone{, and high-end RTX 6000 Ada cards} \parencite{garcia2024step}\hlone{.} \hltwo{Four studies run their method on an embedded device: a Jetson Nano} \parencite{lin2024research}\hltwo{, a Jetson Xavier NX} \parencite{ta2025advanced}\hltwo{, a Jetson AGX Orin} \parencite{zhu2025vertically} \hltwo{and a Google Coral Dev Board} \parencite{ademola2023resource}\hltwo{.}

\hlone{A direct quantitative comparison of accuracy against efficiency across these studies is not meaningful given the current state of reporting, even for the }\hltwo{13 studies that report both an end-to-end accuracy and a frame rate of the whole pipeline}\hlone{. First, the reported accuracies are obtained on private, non-overlapping datasets with substantially different difficulty profiles (controlled gate imagery versus UAV-acquired or multi-perspective in-transit footage). Second, inference speed is highly sensitive to input resolution, and most studies do not state the resolution at which FPS is measured. A model reporting tens of FPS on cropped character patches is not operationally faster than one reporting single-digit FPS on full-HD frames. Third, the hardware platforms differ by orders of magnitude in throughput. We therefore treat the absence of a comparable \textit{accuracy–efficiency tradeoff} as a finding of this review rather than as an analysis we could responsibly produce. Standardised reporting of input resolution, hardware, FPS, parameter count, and computational complexity using an open benchmark is a prerequisite for such comparisons.}

Although the recent studies use the same, established metrics adapted from \hltwo{general-purpose} text detection and recognition \hltwo{tasks}, putting the reported results for \ac{ilu} identification into a frame is not possible with the lack of \hltwo{a widely adopted} benchmark dataset and the \hlone{lack of end-to-end results.} \hltwo{In the 26 studies that report it, the} end-to-end accuracy ranges from \hltwo{19~\%} \parencite{gulsoylu2025trudi} to \hltwo{100~\%} \parencite{kang2026vantage}, \hltwo{with a median of 94.5~\%, a mean of 90.76~\% and a standard deviation of 15.31} percentage points.

\begin{figure}
    \centering
\begin{tikzpicture}
\begin{axis}[
  width=\linewidth, height=0.55\linewidth, font=\small,
  xmin=2000, xmax=2027, ymin=-3, ymax=105,
  xtick={2000,2005,2010,2015,2020,2025},
  x tick label style={/pgf/number format/1000 sep=},
  xlabel={Publication year}, ylabel={End-to-end accuracy / success rate (\%)},
  grid=both, grid style={dotted, black!40},
  legend pos=south west, legend cell align=left, legend style={draw=black!30, font=\footnotesize},
]
\addplot[only marks, mark=*, mark size=2.2pt, mark options={fill=okabeOrange, draw=black, line width=0.3pt}] coordinates {(2023,89.35) (2022,97.3) (2020,96.8) (2016,96.2) (2015,85.8) (2011,91.68) (2004,92.8)};
\addlegendentry{DIP + ML ($n$ = 7)}
\addplot[only marks, mark=*, mark size=2.2pt, mark options={fill=okabeBluishGreen, draw=black, line width=0.3pt}] coordinates {(2024,89.2) (2023,78.4) (2020,93.33) (2019,97.71) (2018,95.41) (2018,93) (2016,91.9) (2016,95) (2025,95)};
\addlegendentry{DIP + DL ($n$ = 9)}
\addplot[only marks, mark=*, mark size=2.2pt, mark options={fill=okabeYellow, draw=black, line width=0.3pt}] coordinates {(2025,96.85) (2025,19) (2024,99.51) (2024,95) (2019,93.98) (2026,96.8) (2025,93.2) (2026,100)};
\addlegendentry{DL ($n$ = 8)}
\addplot[only marks, mark=*, mark size=2.2pt, mark options={fill=okabeVermillion, draw=black, line width=0.3pt}] coordinates {(2023,90.7) (2019,95.92)};
\addlegendentry{DIP + ML + DL ($n$ = 2)}
\addplot[dashed, okabeBlack, mark=square*, mark size=1.8pt] coordinates {(2003,92.8) (2013,88.74) (2018,95.205) (2023,93.2) (2026,98.4)};
\addlegendentry{5-year median}
\addplot[only marks, mark=none, forget plot, nodes near coords, point meta=explicit symbolic, every node near coord/.style={anchor=east, xshift=-3pt, font=\footnotesize}] coordinates {(2025,19) [19~\% (ground, aerial cameras)]};
\end{axis}
\end{tikzpicture}
    \caption{\hlone{End-to-end accuracy versus publication year, with points coloured by method class and a 5-year-bin median overlay} \hltwo{(the last bin covers 2026 only, up to the search cut-off of 12 October 2026)}. \hltwo{The study below 50~\% is annotated with the camera positions of its image acquisition setting.}}
    \label{fig:fig_accuracy_vs_year}
\end{figure}

\hlone{As shown in Figure~\ref{fig:fig_accuracy_vs_year}, the} \hltwo{five-year} \hlone{median is flat across the years}. \hltwo{Similarly, the medians across the method classes are close to each other:} \hlone{DIP + ML }\hltwo{92.8~\% ($n = 7$)}\hlone{, DIP + DL }\hltwo{93.3~\% ($n = 9$)}\hlone{, DL-only }\hltwo{95.9~\% ($n = 8$) and DIP + ML + DL 93.3~\% ($n = 2$)}\hlone{. This does not show that deep learning fails to improve robustness}\hltwo{, nor that it improves it. The values come from different datasets, annotation standards and evaluation protocols. Moreover, the reported end-to-end accuracies describe mainly fixed-camera scenes: 15 use fixed cameras only, 8 do not report the camera position, and only 3 include ground or aerial imagery. For example, classical methods reported their results on} \hlone{narrow, controlled OCR-gate scenes. Pre-2016 papers }\hltwo{($n = 3$)}\hlone{ span }\hltwo{86–93~\%}\hlone{ ($sd = 4$ percentage points), DL-only papers }\hltwo{($n = 8$)}\hlone{ span }\hltwo{19–100~\% ($sd = 27$ percentage points). The wider DL spread is driven by one study that reports end-to-end performance on a genuinely challenging setting (19~\%} \nptextcite{gulsoylu2025trudi}\hlone{). }\hltwo{Excluding it brings the DL median to 96.8~\%; the other seven DL-only studies lie between 93.2 and 100~\%.}

\hlone{These observations motivate two recommendations: (1) end-to-end accuracy should be reported alongside stage-wise metrics whenever possible, and (2) public benchmark datasets with shared evaluation protocols are needed before analysing robustness across operational conditions and method families.}

\hltwo{The reported accuracies are rarely validated in operation. Of the 26 studies that report an end-to-end accuracy, only one describes a system in operational use} \parencite{li2022towards}\hltwo{, and three more report field tests at a terminal} \parencite{kumano2004development, yoon2016automatic, roeksukrungrueang2018implementation}\hltwo{. The remaining end-to-end figures come from offline experiments. Across all 95 studies, six (6.3~\%) describe an operational deployment} \parencite{lee1999automatic, li2022towards, xu2024data, zhao2024practical, cheng2025fast, ta2025advanced}\hltwo{ and seven (7.4~\%) a field test under operating conditions} \parencite{kumano2004development, koo2012novel, yoon2016automatic, roeksukrungrueang2018implementation, wan2021multichannel, gulsoylu2025inland, santos2025automating}\hltwo{. Even these report mainly recognition accuracy and speed. Only one study reports an operational indicator: at a deployed gate, the duration for a truck to pass through the gate area fell from 3--5 minutes to 15--25 seconds} \parencite{ta2025advanced}\hltwo{. None reports gate throughput, the manual-verification rate or cost. The operational benefits attributed to \ac{cv}-based \ac{ilu} identification therefore remain largely undemonstrated, and deployment studies that report such indicators are as important a next step as further accuracy improvements.}

\subsection{Failure Modes}\label{sec:failure_modes}

\hlone{We analysed each of the} \hltwo{95} \hlone{studies for the challenging conditions and failure modes the paper }\hltwo{evaluates against, names as a difficulty of its task}\hlone{ or acknowledges as limitations}. Figure~\ref{fig:fig_failure_modes} \hltwo{summarises the resulting taxonomy of twelve categories (see Supplementary Table~S1, Table~S2 and Table~S3). The twelve categories are not mutually exclusive. A study contributes to every category it reports, so the 656 study--category pairs in Figure~\ref{fig:fig_failure_modes} come from 94 studies, and the percentages, which use $n$ = 95 as the denominator, do not add up to 100~\%. The thirteenth bar, \textit{not reported}, is exclusive with the other twelve.}

\hltwo{The categories were derived from the studies rather than defined in advance. A study counts once for each condition it reports. The detailed rules are given in Supplementary Section~S1. The conditions were first recorded as the studies name them, which yielded 54 conditions (Supplementary Table~S2). They were then grouped into twelve categories by what they concern, for example the illumination of the scene (\textit{lighting}), the physical state of the painted code (\textit{surface condition of the code}) or error sources inside the pipeline (\textit{recognition-level failure causes}).}

\begin{figure}[!h]
    \centering
\begin{tikzpicture}
\begin{axis}[
  scale only axis, width=0.6\linewidth, height=0.546\linewidth, font=\small,
  xbar, bar width=9pt, bar shift=0pt,
  symbolic y coords={{Recognition-level failure causes},{Clutter around the code},{Lighting},{Orientation and viewpoint},{Surface condition of the code},{Character appearance},{Scale},{Layout},{Image quality},{Motion and blur},{Weather},{Occlusion},{Not reported}},
  ytick={{Recognition-level failure causes},{Clutter around the code},{Lighting},{Orientation and viewpoint},{Surface condition of the code},{Character appearance},{Scale},{Layout},{Image quality},{Motion and blur},{Weather},{Occlusion},{Not reported}}, y dir=reverse, enlarge y limits=0.08,
  xmin=0, xmax=100, xtick distance=10,
  xlabel={Number of studies},
  xmajorgrids, grid style={black!15},
]
\addplot[fill=okabeOrange, draw=black, line width=0.3pt] coordinates {(81,{Recognition-level failure causes})};
\addplot[fill=okabeOrange, draw=black, line width=0.3pt] coordinates {(76,{Clutter around the code})};
\addplot[fill=okabeOrange, draw=black, line width=0.3pt] coordinates {(74,{Lighting})};
\addplot[fill=okabeOrange, draw=black, line width=0.3pt] coordinates {(73,{Orientation and viewpoint})};
\addplot[fill=okabeOrange, draw=black, line width=0.3pt] coordinates {(68,{Surface condition of the code})};
\addplot[fill=okabeOrange, draw=black, line width=0.3pt] coordinates {(54,{Character appearance})};
\addplot[fill=okabeOrange, draw=black, line width=0.3pt] coordinates {(52,{Scale})};
\addplot[fill=okabeOrange, draw=black, line width=0.3pt] coordinates {(50,{Layout})};
\addplot[fill=okabeOrange, draw=black, line width=0.3pt] coordinates {(41,{Image quality})};
\addplot[fill=okabeOrange, draw=black, line width=0.3pt] coordinates {(31,{Motion and blur})};
\addplot[fill=okabeOrange, draw=black, line width=0.3pt] coordinates {(30,{Weather})};
\addplot[fill=okabeOrange, draw=black, line width=0.3pt] coordinates {(26,{Occlusion})};
\addplot[fill=neutralGrey, draw=black, line width=0.3pt] coordinates {(1,{Not reported})};
\node[anchor=west, xshift=2pt, font=\footnotesize] at (axis cs:81,{Recognition-level failure causes}) {81 (85~\%)};
\node[anchor=west, xshift=2pt, font=\footnotesize] at (axis cs:76,{Clutter around the code}) {76 (80~\%)};
\node[anchor=west, xshift=2pt, font=\footnotesize] at (axis cs:74,{Lighting}) {74 (78~\%)};
\node[anchor=west, xshift=2pt, font=\footnotesize] at (axis cs:73,{Orientation and viewpoint}) {73 (77~\%)};
\node[anchor=west, xshift=2pt, font=\footnotesize] at (axis cs:68,{Surface condition of the code}) {68 (72~\%)};
\node[anchor=west, xshift=2pt, font=\footnotesize] at (axis cs:54,{Character appearance}) {54 (57~\%)};
\node[anchor=west, xshift=2pt, font=\footnotesize] at (axis cs:52,{Scale}) {52 (55~\%)};
\node[anchor=west, xshift=2pt, font=\footnotesize] at (axis cs:50,{Layout}) {50 (53~\%)};
\node[anchor=west, xshift=2pt, font=\footnotesize] at (axis cs:41,{Image quality}) {41 (43~\%)};
\node[anchor=west, xshift=2pt, font=\footnotesize] at (axis cs:31,{Motion and blur}) {31 (33~\%)};
\node[anchor=west, xshift=2pt, font=\footnotesize] at (axis cs:30,{Weather}) {30 (32~\%)};
\node[anchor=west, xshift=2pt, font=\footnotesize] at (axis cs:26,{Occlusion}) {26 (27~\%)};
\node[anchor=west, xshift=2pt, font=\footnotesize] at (axis cs:1,{Not reported}) {1 (1~\%)};
\end{axis}
\end{tikzpicture}
    \caption{Reported failure modes and challenging conditions. A study \hltwo{may appear in several bars if it reports conditions of several categories. Percentages are relative to} \hltwo{$n = 95$}.}
    \label{fig:fig_failure_modes}
\end{figure}

\hltwo{Recognition-level failure causes are the most frequently reported category (85~\% of studies), followed by clutter around the code (80~\%), lighting (78~\%), orientation and viewpoint (77~\%), the surface condition of the code (72~\%), character appearance (57~\%), scale (55~\%), layout (53~\%) and image quality (43~\%). Motion and blur (33~\%), weather (32~\%) and occlusion (27~\%) are reported least. Recognition-level failure causes lead partly because the category is broad. It collects similar-looking or touching characters, segmentation, detection and format errors, and acquisition failures. The single conditions reported most often are illumination variation (58 studies), detection or localisation errors (57), vertical codes (52) and damaged or deformed characters (47).} \hlone{Failure modes characteristic of mobile deployments, }\hltwo{motion blur and occlusion, are therefore}\hlone{ notably under-represented. This imbalance is best understood as a research gap created by the }\hltwo{static OCR-gate origin}\hlone{ of the field.}

\hltwo{Only 1 of 95 studies (1~\%) reports no challenging condition or failure mode at all, and the median study names 12 conditions. Reporting a condition, however, does not mean that the study reports results for it. A condition also counts when a study names it as a difficulty or when it is present in the study's data, without results broken down by condition.} \hlone{Combined with the dataset-availability gap, this prevents meaningful cross-study comparison of robustness claims.}

\subsection{Future Research Directions}\label{sec:future_directions}
\hltwo{Five directions follow from this review for moving identification beyond the gate: a public benchmark, robustness to occlusion and motion blur, the partition of computation between vehicle and server, pose estimation of the \acp{ilu}, and mission planning for the vehicles that carry the cameras. They are discussed below together with the directions proposed by the reviewed studies.}

In the literature, future directions for the \ac{ilu} identification task mainly focus on enhancing the datasets by including more images taken from diverse environments, introducing different challenges such as various distances, angles, weather and lighting conditions\hltwo{. 17 studies propose larger or more diverse datasets} \parencite[e.g.,][]{liu2025lightweight, yu2024two, lau2024modelling, diaz2024wagon, safitri2026vgg16, santos2025automating}. This expansion aims to cover multiple ports and scenarios to improve model generalisation. However, increasing the diversity of proprietary datasets would not solve the fundamental problem with the datasets for \ac{ilu} identification\hltwo{: most of them are not public, and currently, none serves as a common benchmark} (Section~\ref{sec:datasets}). \hltwo{Therefore, a public benchmark is the first priority: images from fixed, ground and aerial cameras that cover all \ac{ilu} types and are labelled per condition and per pipeline stage, so that end-to-end accuracy and stage-wise metrics can be reported on the same data.} Given that creating a new dataset for downstream \ac{cv} tasks is tedious, time consuming, and costly \parencite{wu2023diffumask, tejero2023full}, alternatives such as synthetic data generation can be considered \parencite{guan2024bridging, gupta2016synthetic}, \hltwo{e.g., by using a game engine} \parencite{long2020unrealtext}\hltwo{. Six reviewed studies have taken first steps with synthetic or generated training images} \parencite{verma2016automatic, liang2020research, xu2024data, lau2024modelling, lin2024research, cheng2025fast}\hltwo{, and three simulate challenging conditions to diversify their data} \parencite{yang2023lightweight, zhao2021improved, lin2024research}\hltwo{. Synthetic scenes that vary viewpoint, distance, occlusion and weather together, as vehicle-mounted cameras require, remain to be explored.

Class balance is equally important. The reviewed datasets are almost exclusively container datasets (all 95 studies use container images, four also include semi-trailers and none treats swap bodies as a class of its own) and no study reports class-wise results, although swap bodies and semi-trailers dominate European rail--road terminals (EN 13044} \nptextcite{UIRR2011new}\hltwo{). Future datasets should cover all \ac{ilu} classes and report performance per class.}

Handling challenging conditions, such as low-light scenarios, occlusions, and degraded or damaged ID codes, is another direction pointed by the researchers. \hltwo{For vehicle-mounted cameras, occlusion and motion blur matter most, yet they are among the least reported conditions} (Section~\ref{sec:failure_modes})\hltwo{; test sets labelled per condition would show how well methods cope with them.} Suggested methods include image enhancement techniques such as noise reduction \parencite{hlabisa2024automated}, \hltwo{the repair of the damaged code text before recognition} \parencite{lin2024research}\hltwo{,} or the use of \acused{gan}\aclp{gan} (\acsp{gan}; \nptextcite{goodfellow2014generative, radford2015unsupervised}) to address issues such as shadows, rust, and dirt \parencite{khuboni2025confidence}\hltwo{. Low-light enhancement} \parencite{lin2024research} \hltwo{and deep denoising} \parencite{ta2025advanced} \hltwo{have been integrated into \ac{ilu} pipelines, which makes a systematic study of their effect under each condition a promising next step.}

\hlone{Super-resolution} \parencite{dong2015image} \hlone{is a task that can tackle the problems with degraded or damaged ID codes}\hltwo{. A first study applies a generic super-resolution model to damaged codes} \parencite{lin2024research}\hlone{. However, the} literature on scene-text super-resolution has focused primarily on natural language, leaving an opportunity for research specifically on contextless text such as \ac{ilu} codes. Alternatively, multi-view \ac{ilu} identification is another future direction due to its potential against degraded ID codes \parencite{khuboni2025confidence, chen2011hidden, hlabisa2024automated, wu2012automated, kumano2004development, samsudin2026freight} as multiple images of a single \ac{ilu} are captured from different angles in this approach\hltwo{. Fusing several views} \parencite{yoon2016automatic, mei2016novel, zhiming2019automatic, shithil2022container} \hltwo{or several frames of a video} \parencite{wu2019key, hsu2023automatic, geirinhas2025divide} \hltwo{has so far been tried at fixed installations. Vehicle-mounted cameras observe an \ac{ilu} from changing viewpoints by design, so these approaches deserve further exploration in that setting, as several studies also propose} \parencite{gulsoylu2025trudi, wu2026container, al20168, ta2025advanced}\hltwo{.}  Post-processing steps are also recommended to verify recognition results by using the check digit and the owner code \parencite{verma2016automatic, khuboni2025confidence, he2005new, wang2017container, nguyen2020automatic}\hltwo{. The check-digit test is common} \parencite[e.g.,][]{yoon2016automatic, li2022towards, zhiming2019automatic, lin2024research, diaz2024wagon, santos2025automating}\hltwo{, and a few studies also check the owner code or the code format} \parencite{kumano2004development, roumpattana2026enhancing, gulsoylu2025inland}.

For \ac{dl}-based methods, \hltwo{one suggestion is to refine} network architectures to increase the receptive field so that the model can \textit{see} more of the scene, especially smaller objects\hltwo{,} in a single pass \parencite{liu2025lightweight, feng2020improved}. Another future work suggestion is incorporating attention mechanisms to boost recognition accuracy \parencite{feng2020port, khuboni2025confidence} \hlone{or optimising the tokenisation process of vision transformer-based architectures as there are always the same amount of characters expected for} \hltwo{ISO~6346}-compliant \hlone{ID codes.} \hltwo{Attention modules are part of several recent pipelines} \parencite[e.g.,][]{zhang2021vertical, lin2024research, yao2025advanced, ta2025advanced}\hltwo{, whereas adapting the tokenisation to the fixed code length has not been explored yet.} \hlone{For better applicability}, the development of unified, end-to-end models is suggested to streamline processing pipelines \parencite{xu2024data, lin2024research, li2019automatic, wan2021multichannel, zhu2025vertically, wu2026container}\hltwo{. Five such models among the reviewed studies} \parencite{zhang2021vertical, zhao2024practical, garcia2024step, chen2026end, liang2025dimension} \hltwo{open this direction.} \hlone{While \ac{dl} models achieve high accuracy, their computational demands shape where in the pipeline computation should be spent rather than mandating fully on-device inference.

For vehicle-mounted setups, an edge-to-cloud offloading architecture is a concrete and domain-specific direction. A lightweight on-vehicle model can perform frame selection }\hltwo{(discarding occluded, blurred or code-absent frames using inexpensive image-quality cues such as a blur metric or a text-presence score)}\hlone{ and }\hltwo{forward}\hlone{ only candidate frames to a central server for full text spotting. This reframes the \textit{accuracy vs.~efficiency} question as an IoT and system-design choice (what runs where, with what bandwidth budget) rather than a single-model tradeoff. Complementary directions include model compression, quantisation, and neural architecture search for the edge-side selector, as well as adaptive frame-rate policies tied to vehicle speed and proximity to} \acp{ilu}\hltwo{. Building blocks for this exist at fixed installations: frame selection} \parencite{wu2019key, tang2020container, geirinhas2025divide}\hltwo{, quantisation} \parencite{ademola2023resource}\hltwo{, knowledge distillation} \parencite{kang2026vantage} \hltwo{and edge inference at a deployed gate, with a central server for training and display} \parencite{ta2025advanced}\hltwo{. Further studies call for real-time or edge deployment} \parencite{yu2024two, zhang2021vertical, wu2026container, tran2026automatic, roumpattana2026enhancing}\hltwo{. Combining these into an offloading architecture for vehicle-mounted cameras remains open.}

The contextless nature of \hltwo{ISO~6346} codes makes scene-text spotting task harder as there are no natural language cues \parencite{fang2021read}. Therefore, directly applying current scene-text recognition models that rely on semantic context, is not the most suitable approach, as they usually use a language model or at least a decoder module pre-trained on natural language to correct the initial recognition result \parencite{zhang2025linguistics, li2023trocr}. However, incorporating complex language models \parencite{fang2021read} increases model complexity\hltwo{, while their benefit for contextless ID codes has not been confirmed (Section~\ref{sec:terminology_beyon_ocr_gates}). Conditioning the recognition on the code format instead, for example with regular expressions, is a promising alternative} \parencite{garcia2024step}\hltwo{.} \hltwo{First studies take the contextless nature of the codes into account, by choosing recognisers without a language model} \parencite{gulsoylu2025trudi, cheng2025fast, wu2026container} \hltwo{or by encoding the code format} \parencite{garcia2024step, liang2025dimension}\hltwo{. These first steps make research on contextless scene-text recognition models} \hltwo{a promising} opportunity which may result in novel and more efficient methodologies \hlone{that are better suited for transportation and }\hltwo{logistics applications}. These new methods can also be useful for similar \hlone{transportation related} applications including \ac{uic}-wagon codes \parencite{diaz2024wagon, garcia2024step}, or registration plates \parencite{ta2025advanced}\hltwo{, as several reviewed studies suggest} \parencite{yoon2016automatic, wu2015new, chen2011hidden, roumpattana2026enhancing}\hltwo{.}

\hlone{Moreover, we suggest other directions, highly relevant for terminal operations. Once cameras are mounted on multiple terminal vehicles, and CCTV cameras are included as a source, the same} \hltwo{\ac{ilu}} \hlone{is observed by a sequence of cameras (e.g., gate → straddle carrier → quay crane) and identification becomes a tracking problem across heterogeneous viewpoints rather than a sequence of independent recognition events. Therefore, }\hltwo{approaching the ILU identification task as} \hlone{a tracking problem would support the port monitoring holistically. The actionable direction is to fuse ILU identification with the Terminal Operating System's knowledge. This could be done by using existing inventory in the terminal, and ground vehicles' own task schedule as priors.} \hltwo{Several reviewed studies also call for integration with port systems} \parencite{yao2025advanced, samsudin2026freight, ta2025advanced, santos2025automating, sharma2013smart, sharma2013novel}\hltwo{, and one outlines it} \parencite{liang2025dimension}\hltwo{. Implementing and evaluating such an integration is still open.}

\hlone{Another unexplored area is the uncertainty calibration of the ILU identification.} \hltwo{Confidence scores are already used to select segments, recognisers or frames or to flag uncertain reads} \parencite{chen2011hidden, zhang2021vertical, nguyen2023digital, khuboni2025confidence, geirinhas2025divide}\hltwo{, but calibrating them is the next step.} \hlone{ A terminal operator does not need a CV system that is always right; they need one that knows when it is wrong. An identification reported with calibrated uncertainty enables a clean operational pattern: high-confidence identifications proceed automatically, low-confidence identifications route to a manual review queue that the operator works through asynchronously, and only the small middle band requires immediate human attention. Reporting confidence as the softmax of the final classification layer does not constitute calibration. The model's confidence must correspond to its empirical accuracy. This }\hltwo{can be achieved, for example, with}\hlone{ temperature scaling} \parencite{guo2017calibration}\hlone{, deep ensembles} \parencite{lakshminarayanan2017simple} \hlone{or evidential learning} \parencite{sensoy2018evidential}\hlone{, all of which must be validated on a publicly available test set. Future work should treat calibrated, per-identification uncertainty as an equally important deliverable alongside end-to-end accuracy.}

\hltwo{Vehicle-mounted cameras also make two neighbouring tasks part of identification. Pose estimation, predicting the real-world position and orientation of each \ac{ilu}, links an identified code to a slot in the yard and to the records of the Terminal Operating System.

Unmanned vehicles need an automatically planned coverage path that brings the code of every \ac{ilu} into view at a distance and angle at which it can be recognised. Planning the coverage path together with the identification method remains open.}

\hltwo{Finally, as shown in Section~\ref{sec:evaluation}, operational deployments are rare in the literature and only one of them reports an operational indicator, the gate passage time per truck. Deployment studies that measure gate throughput, truck turnaround time, the manual-verification rate and cost before and after the introduction of \ac{cv}-based identification are needed to substantiate the benefits attributed to it, as several reviewed studies also propose} \parencite{santos2025automating, zhao2024practical, ngo2023vision, shithil2022container, wu2026container}\hltwo{.}

\subsection{\texorpdfstring{\protect\hltwo{Limitations}}{Limitations}}\label{sec:limitations}
\hltwo{This review has several limitations. First, Google Scholar ranks its results by relevance, serves a limited number of results per string and allows no reproducible export. Second, only articles in English were included, and theses, project reports and patents were excluded (Table~\ref{tab:eligibility}). Studies from non-English-speaking countries may be under-represented, and commercial systems, such as most installed \ac{ocr}-gates, are included only where their developers published them. Third, 2026 is covered only up to the cut-off of 12 October, so its five studies represent only part of that year. Fourth, the synthesis depends on what the studies report. 39 of the 95 studies (41~\%) do not report the camera position, only 26 report an end-to-end accuracy for the whole pipeline on the whole test set, and most evaluate on private datasets. The analyses of camera positions and end-to-end accuracies therefore rest on subsets of the studies, results on private datasets cannot be compared directly, and a failure mode that a study does not mention may still occur in its data. As is usual for scoping reviews, the methodological quality of the studies was not appraised, and their results are reported as published} \parencite{tricco2018prisma}.

\section{Conclusion}
\hltwo{In recent years, the number of publications has grown, and so has the share published in ranked outlets (from 23~\% in 2016--2020 to 64~\% in 2021--2025).} Addressing this growth, this review provides the first systematic synthesis of \ac{cv}-based \ac{ilu} identification research by analysing \hltwo{95} empirical studies published over three decades, \hltwo{and proposes a more inclusive terminology}.

\hltwo{The review answers its four research questions as follows. (RQ1) The literature mostly speaks of \textit{container code} or \textit{container number} recognition, terms that cover neither semi-trailers and swap bodies nor the detection step. We therefore propose the term \textit{\ac{ilu} identification}. (RQ2) Until 2016, only classical image processing and machine learning were used. Since then, deep learning has dominated, driven by GPU hardware and open-source frameworks, and 13 of the 14 studies that use transformers appeared in 2024--2026. (RQ3) The datasets of most studies are private (82~\%), only eight datasets are publicly available, and only a small portion of the studies report an end-to-end accuracy. Therefore, the published results can hardly be compared. (RQ4) The conditions typical of vehicle-mounted cameras, occlusion and motion blur, are among the least reported. Only six studies describe a deployed system and seven a field test. The most relevant directions are a public benchmark, robustness to occlusion and motion blur, offloading computation from edge devices to servers, pose estimation of the \acp{ilu} and mission planning} (Section~\ref{sec:future_directions}).

Over the years, the field has shifted from classical \ac{dip} to \ac{dl} thanks to the advancements in computational resources and deep learning frameworks. This shift comes with the need \hltwo{for large} datasets to train \ac{dl} models\hltwo{,} \hlone{and dataset accessibility is} \hltwo{a major} obstacle\hltwo{,} with \hltwo{the datasets of 82.11~\% of the studies} being private. \hlone{This creates problems for both researchers} and \hlone{practitioners. While large seaports have the means of creating large scale datasets, small sized ports have limitations to create datasets with the same scale, which }\hltwo{may create}\hlone{ a competitive disadvantage}. The median dataset size \hltwo{of the deep learning studies (3,000 images)} is \hltwo{small for} \ac{dl} training\hltwo{, which leads} researchers to adapt \hltwo{general-purpose} text detection and recognition models\hltwo{, whose natural-language priors do not fit codes without semantic meaning}. \hlone{Limited dataset accessibility also affects} the quality and credibility of publications as there are no \hltwo{adopted} benchmark datasets. This makes performance comparisons difficult \hlone{along with the lack of end-to-end analysis}, leaving gaps in understanding which methods perform best under which setting. To the best of our knowledge, there are only \hltwo{eight} \hlone{publicly available datasets which can be adopted for benchmarking. Considering that} \hltwo{three of the four released by reviewed studies appeared in 2024 or 2025}, \hlone{in the near future}, \hltwo{the adoption of these datasets} might \hltwo{ease} this problem.

\hlone{While the rise of \ac{dl}-based methods brought the need for} \hltwo{publicly available datasets, the shift} \hlone{from static to dynamic image acquisition settings brings other dimensions}. \hltwo{Those datasets should also cover swap bodies and semi-trailers, which no reviewed study evaluates separately.} The research on \ac{ilu} identification relied on fixed camera settings such as \ac{ocr}-gates. \hlone{However, recent studies are increasingly adopting }\hltwo{vehicle-mounted}\hlone{ camera settings to broaden operational coverage of the yard}\hltwo{. The systems are also becoming more connected. Instead of one device serving one gate, cameras on several unmanned vehicles may in the future collect images that are processed partly on edge devices and partly on central servers.}\hlone{ This shift introduces new challenges like higher variations in object scale caused by changing camera-to-object distances, unknown \ac{ilu} locations, and the system-design question of how to partition computation between the vehicle and a central server}\hltwo{, a question that arises even when the identification relies on \ac{cv} alone}\hlone{.} \hltwo{This is why other \ac{cv}, robotics and IoT related tasks, such as pose estimation (predicting the real-world coordinates and orientation of \acp{ilu}), mission planning for autonomous vehicles (automatic coverage path planning, so that the code of every \ac{ilu} comes into view of a camera) and edge-to-cloud offloading, are now emerging priorities.} \hlone{The seemingly simple problem of identifying \acp{ilu} is} a \hlone{fundamental step for autonomous terminal operations}. \hltwo{Therefore, addressing these challenges holds potential for transportation and logistics. For example, precise monitoring could optimise terminal operations, reduce human errors and pave the way for faster turnaround times, lower operational costs and increased resilience against disruptions such as }\hltwo{misplaced \acp{ilu}}. \hltwo{Whether these benefits materialise remains to be shown in deployment studies.}

\section*{Funding}
This work was supported by the German Federal Ministry for Digital and Transport (BMDV) through the funding programme Innovative Hafentechnologien II (IHATEC II), grant number 19H23002C. The funder had no role in the design of the study, in the collection, analysis or interpretation of the data, in the writing of the manuscript or in the decision to submit the article for publication.

\section*{Declaration of the use of AI}
During the preparation of this work the authors used Claude, and Mistral as AI assistants, in order to (a)~revise the figures, and tables, (b)~write a browser based the retrieval and screening tool, (c)~review the extracted data to cover human errors by checking the chart against the full texts, (d)~review the manuscript and proofread it, and (e)~draft the highlights and the statements on this title page. The authors reviewed and edited the content as needed and take full responsibility for the content of the publication.

\section*{Data availability}
The supplementary file \texttt{ilu\_identification\_data\_chart.xlsx} holds the data chart of this review, with one row per included study and verbatim quotes from the studies as evidence for every extracted item. All reviewed studies are published and cited in the article or its supplementary material.

\printbibliography

\end{document}